\documentclass{article}

\usepackage{arxiv}

\usepackage[utf8]{inputenc} 
\usepackage[T1]{fontenc}    
\usepackage{hyperref}       
\usepackage{url}            
\usepackage{booktabs}       
\usepackage{amsmath}
\usepackage{amsfonts}       
\usepackage{nicefrac}       
\usepackage{microtype}      
\usepackage{graphicx}
\graphicspath{ {./images/} }
\usepackage{bbm}
\usepackage{array}
\usepackage[ruled, vlined]{algorithm2e}
\usepackage{color}
\usepackage{siunitx}

\usepackage{colortbl}
\newcolumntype{P}[1]{>{\centering\arraybackslash}p{#1}}
\newcolumntype{R}[1]{>{\raggedleft\arraybackslash}p{#1}}

\usepackage{tikz}
\usetikzlibrary{positioning,fit,backgrounds,arrows.meta,calc,matrix,decorations.pathreplacing}

\definecolor{c_diag}{rgb}{0.85,0.89,0.953}
\definecolor{t_diag}{rgb}{0,0,0}
\definecolor{c_ndiag}{rgb}{.95,0.95,0.95}

\definecolor{c_hlight}{rgb}{.886,0.9411,0.851}
\definecolor{c_hdark}{rgb}{.663,0.82,0.557}
\definecolor{c_rest}{rgb}{.984,0.898,0.839}

\usepackage{natbib}

\newlength{\AffCellW}
\newcommand{\MaxAffinity}{0.375}

\newcolumntype{L}{>{\raggedright\arraybackslash}m{1.25cm}}
\newcolumntype{C}{>{\centering\arraybackslash}m{1.18cm}}

\newcommand{\rowhdr}[1]{\centering\bfseries #1}

\newcommand{\databar}[1]{%
    \begin{tikzpicture}[baseline=(txt.base)]
        \pgfmathsetmacro{\ratio}{min(max(#1/\MaxAffinity,0),1)}
        \pgfmathsetlengthmacro{\barw}{\ratio*\AffCellW}

        \path[use as bounding box] (0,0) rectangle (\AffCellW,2.8ex);

        \ifdim\barw>0pt
            \shade[left color=green!20,right color=green!55]
                (0,0.2ex) rectangle (\barw,2.45ex);
        \fi

        \node[anchor=base] (txt) at (.5\AffCellW,0.75ex) {\strut \num{#1}};
    \end{tikzpicture}%
}

\title{ITL: Interpretable Document Alignment with Structured Reference Frameworks}

\author{
 Raúl Giráldez\\
  School of Engineering\\ Pablo de Olavide University\\ 
  ES-41013 Seville, Spain \\
  \texttt{giraldez@upo.es} \\
   \And
Dayrelis Mena \\
  School of Engineering\\ Pablo de Olavide University\\ 
  ES-41013 Seville, Spain \\
  \texttt{dmentor@upo.es} \\
   \And
 Jes\'us S. Aguilar--Ruiz \\
  School of Engineering\\ Pablo de Olavide University\\ 
  ES-41013 Seville, Spain \\
  \texttt{aguilar@upo.es} \\
}

\begin{document}
\maketitle
\begin{abstract}
Measuring alignment between documents and structured reference frameworks requires identifying conceptual evidence distributed throughout the text and reporting it through measures that are quantitative, interpretable, and traceable. Many commonly used retrieval and classification approaches return either pairwise similarity scores or one or more class labels, whereas fewer methods provide concept-level scores that are directly traceable to the terminological evidence supporting them. We present \emph{Intelligent Target Locator} (ITL), a domain-agnostic and language-portable methodology that estimates the affinity between the textual units of a target document and the concepts defined in a \emph{Structured Reference Document} ($SRD$). From the $SRD$, ITL induces concept-specific terminological profiles built from independent terms, bigrams, trigrams, and co-occurrences. Each term is assigned an importance weight that combines concept membership, term-type specificity and inter-concept discriminability. The output is a textual-unit--concept affinity matrix that can be aggregated at different levels of granularity. We conduct an internal consistency assessment using the 17 Sustainable Development Goals (SDGs), evaluating each official goal statement against the $SRD$ induced from the same set of descriptors. Every statement reached its highest affinity with the corresponding concept, and the mean affinity across the remaining concepts stayed marginal relative to the mean reference affinity. This separation indicates that ITL distinguishes the conceptual profiles of the framework. ITL thus offers a general basis for quantifying document alignment with structured frameworks while keeping each result traceable to the terminological evidence that supports it.
\end{abstract}


\section{Introduction}\label{sec:intro}

Measuring how closely a document aligns with a structured reference framework is a recurrent task in knowledge management and in assessing compliance with regulatory, strategic, or program frameworks. Many organisations ---including public authorities, universities, businesses and non-governmental organisations---need to assess consistently and traceably the extent to which their documentation aligns with a reference framework. This assessment is complicated by the growing volume and heterogeneity of such documentation, its frequent availability in unstructured formats such as PDF, and the semantic ambiguity of high-level texts. Manual assessment is therefore costly, hard to reproduce, and dependent on expert judgment.

Operationalizing document \emph{alignment} is not straightforward. The relationship between a document and a reference framework is primarily conceptual rather than strictly lexical, and the evidence for it may be scattered across different parts of the text. Identifying which elements of the framework relate to the document is often not enough: in many settings the relationship must also be quantified through a measure that is interpretable, auditable, and comparable across documents, and that supports fine-grained analysis and explanation of the alignment found.

We approach this problem with a general formulation that is independent of domain and language. We estimate the relationship between a target document $D$ and a \emph{Structured Reference Document} ($SRD$), which models a domain as a finite set of concepts, each described by a reference textual unit. The $SRD$ thus acts as the terminological reference against which the alignment of external documents is assessed.

Within this formulation we present ITL (\emph{Intelligent Target Locator}), a computational model that builds an interpretable profile of a document $D$ against the concepts of the $SRD$ most closely related to it and quantifies how strongly those concepts are represented in $D$, using an original metric we term \emph{affinity}. Affinity is a weighted measure of the extent to which $D$ shares relevant concepts with the $SRD$, taking into account their importance within the reference framework. Unlike a single-label classifier, ITL computes an affinity value for every (textual unit, concept) pair, so a single textual unit may align with several concepts when its content warrants it. This design allows concepts to be ranked by relevance and affinity to be analyzed at different levels of granularity within $D$ (the whole document, paragraphs, or sentences), as well as within the $SRD$.

The contribution of this research is twofold. First, we formalize the affinity metric, which is computed from the weighted number of related concepts shared by $D$ and the $SRD$, together with the importance of those concepts within the reference framework. Second, we propose a general two-phase \emph{methodology}. During the reference-model induction phase, a structured and attributed model is induced from the $SRD$ to aggregate and weight its concepts. During the document-alignment phase, a document $D$ is processed to extract and structure its concepts, after which its affinity with each concept in the $SRD$ is computed. Both phases are therefore fully determined by the provided $SRD$ and do not depend on a specific domain or language.

The original motivation for this work arose within the \emph{Universities for Sustainable Development}\footnote{\textit{Universities for Sustainable Development} (USD) was an Erasmus+ project funded by the European Commission under the KA220-HED action (Cooperation Partnerships in Higher Education), reference 2021-1-ES01-KA220-HED-000029950 (2022--2025), with the participation of Pablo de Olavide University. Further information: \url{https://www.upo.es/usd-project/}. } project, which focused on the Sustainable Development Goals (SDGs) of the United Nations 2030 Agenda~\cite{UN2030Agenda2015} in the context of European universities. We use the SDGs as a consistency study and as the basis for the empirical validation of the method. Specifically, we instantiate the $SRD$ using the official text of the goals and evaluate the ability of ITL to assign each textual unit to its corresponding concept. The procedure is not restricted to this setting, however, and can be applied to any context in which a reference document is organized into concepts, including regulations, standards, strategies, and regulatory frameworks.

The remainder of the paper is organized as follows. Section~\ref{sec:related} reviews the state of the art and positions the proposed approach. Section~\ref{sec:problem} formalizes the problem, notation, and the affinity metric. Section~\ref{sec:methodology}  describes the ITL methodology and its reference-model induction and document-alignment phases. Section~\ref{sec:case} presents the SDG consistency study and the empirical validation of the method. Finally, Section~\ref{sec:conclusions} presents the conclusions and directions for future work.

\section{Related Work}\label{sec:related}

The problem of estimating affinity between a document and a concept-structured reference framework intersects several research directions: (i) semantic similarity and information retrieval for document comparison; (ii) concept and terminology extraction and topic modeling for comparable content representation; (iii) formal representation of reference frameworks through ontologies and knowledge graphs; and (iv) automatic text classification against predefined thematic frameworks. In these contexts, the Sustainable Development Goals (SDGs) are a representative application domain. The following subsections review these research areas and position ITL with respect to the methodological gap addressed in this work.

\subsection{Semantic Similarity and Information Retrieval}

The comparison of documents with reference repositories has traditionally been addressed within Information Retrieval. Lexical models such as BM25 remain competitive due to their efficiency and robustness in general retrieval settings~\citep{RobertsonZaragoza2009}. However, approaches based on term matching are limited when semantic variation is present, including synonymy and paraphrasing, and do not directly capture relationships at the conceptual level.

The development of dense representations has shifted part of the research focus toward semantic approaches. Transformer-based models such as BERT generate contextual \emph{embeddings} for text matching and classification tasks~\citep{Devlin2019BERT}, while Siamese architectures such as Sentence-BERT enable efficient computation of sentence-level similarity~\citep{Reimers2019SBERT}. For large-scale retrieval and alignment, dense retrieval methods such as DPR~\citep{Karpukhin2020DPR} and late-interaction architectures such as ColBERT~\citep{KhattabZaharia2020ColBERT} have also been proposed.

Dense retrieval models provide strong semantic matching capabilities, although their native retrieval scores do not generally decompose a match into concept-specific terminological evidence. Attribution or explainability methods may be applied subsequently, whereas ITL incorporates term-level traceability directly into its scoring rule. Instead, ITL quantifies the affinity with respect to \emph{each} concept in $SRD$ and preserves traceability to the terms contributing to the resulting value. The two approaches are complementary, since dense representations could be incorporated as a matching component within the ITL procedure.

\subsection{Concept Extraction, Terminology, and Topic Modeling}

Affinity estimation requires the document and reference framework to be represented in comparable forms. The extraction of concepts and terminology therefore provides a natural basis for this task. Linguistic methods based on syntactic patterns have been proposed for terminology recognition~\citep{JustesonKatz1995TechnicalTerminology}, together with hybrid linguistic-statistical approaches for multi-word terms, such as the C-value/NC-value method~\citep{Frantzi2000CValue}. Keyword extraction has also been addressed through unsupervised graph-based methods such as TextRank~\citep{MihalceaTarau2004TextRank} and lightweight document-level methods that do not require a training corpus, such as YAKE!~\citep{Campos2020YAKE}. Topic-modeling methods, particularly LDA~\citep{Blei2003LDA}, have been used to identify latent structure in document collections and support thematic comparison.

ITL draws on this research direction through a linguistic-processing core that extracts four term types (independent terms, bigrams, trigrams, and co-occurrences) to characterize each concept in the $SRD$. Unlike topic modeling, which induces latent topics without explicit correspondence to a predefined framework, ITL anchors the representation to the explicit concepts defined in the $SRD$ and weights each term according to its discriminability and specificity. This representation provides the basis for an interpretable affinity metric.

\subsection{Formal Representation of Reference Frameworks: Ontologies and Knowledge Graphs}

The formal representation of reference frameworks has been extensively addressed within the Semantic Web and Ontology Engineering communities. In the SDG domain, knowledge organization systems have been developed to support interoperability and linkage with open data~\citep{Joshi2021SDGKOS}, together with knowledge graphs for progress monitoring, such as SustainGraph~\citep{Fotopoulou2022SustainGraph}.

These approaches primarily address the representation and linkage of entities and data rather than the comparison of external documents with a reference framework and the quantification of their alignment. ITL tackles a complementary problem. It does not define a new ontology; instead, it takes a reference document as the $SRD$, which may originate from resources such as those described above or directly from textual concept definitions, and defines an affinity measure for evaluating external documents against that reference.

\subsection{Automatic Classification against Thematic Frameworks: The SDG Case}

A specific instance of document-to-framework comparison is the automatic assignment of text to categories defined by a thematic framework. The SDGs have been widely used as an application domain for this problem. Multi-label classification tools based on deep learning have been proposed, including SDG-Meter~\citep{Guisiano2022SDGMeter} and OSDG~\citep{Pukelis2022OSDG}, together with multilingual datasets for evaluation, such as the SDGi Corpus~\citep{Skrynnyk2024SDGiCorpus}.

These approaches address a \emph{labeling problem}: they assign one or more categories to a text, frequently using supervised classifiers that require annotated data and produce labels or probabilities without explicitly decomposing the result into the evidence supporting the assignment. ITL targets a different and more general problem. Rather than classifying text against a fixed set of learned categories, it computes a graded affinity with respect to each concept defined in an arbitrary $SRD$. The resulting affinities are interpretable, traceable, and aggregable at different levels, and the reference framework can be changed without retraining a classifier. In this work, the SDGs constitute one such framework and are used as the validation case; the methodology itself is not specific to this domain.

\subsection{Positioning of ITL}

ITL is defined by three properties that, taken together, are not provided by any single line of work reviewed above. First, its reference model is induced directly from a concept-structured reference document (the $SRD$), rather than fitted to a fixed, pre-learned category set: the $SRD$ itself supplies the terminological profiles against which alignment is assessed. Second, the resulting alignment is expressed as a concept-level, graded affinity for every (textual unit, concept) pair, rather than a single similarity score or a discrete label, supporting multi-concept association and analysis at different levels of granularity. Third, each affinity value is intrinsically traceable to the specific terms of the target document and their weighted contribution, without requiring a separate explainability layer.

None of the reviewed research directions provides these three properties jointly. Dense retrieval and similarity methods offer strong matching capabilities but no native decomposition into concept-specific terminological evidence. Concept- and terminology-extraction methods supply comparable representations but no mechanism for inducing a reference model tied to a predefined concept structure. Ontological and knowledge-graph resources formalize reference frameworks but are not designed to compute a graded alignment measure against them. Automatic classification methods assign discrete labels from learned categories, typically require annotated training data, and offer limited traceability to the underlying evidence.

\section{Problem Formalization}\label{sec:problem}

\subsection{Problem Statement}
\label{subsec:problem-statement}

Let $SRD$ denote a \emph{Structured Reference Document} that encodes a target semantic domain through a finite set of concepts $\mathcal{C}=\{c_1,\dots,c_m\}$. Each concept $c_j$ is described by a reference textual unit $R_j$ (e.g., a definition, paragraph, or text fragment), such that the set of units in the $SRD$ is represented as $U(SRD)=\{R_1,\dots,R_m\}$. The $SRD$ therefore serves as a semantic and terminological reference against which the alignment of external documents is assessed.

Let $D$ denote a \emph{target document} composed of an ordered sequence of textual units $U(D)=\{S_1,\dots,S_n\}$ (e.g., sentences, segments, or paragraphs) obtained through a segmentation process. Hereafter, $S_i$ denotes the $i$-th textual unit of document $D$, with $i\in\{1,\dots,n\}$. Analogously, $SRD$ is represented through its reference textual units $R_j$, each associated with a concept $c_j$, with $j\in\{1,\dots,m\}$.

The objective is to quantify the degree of alignment between each unit $S_i$ in $D$ and each concept $c_j$ in the $SRD$. This makes it possible to identify semantic relationships at the textual-unit level while supporting subsequent aggregation at the paragraph, section, and complete-document levels. Consequently, the resulting representation supports interpretable analyses and visualization techniques, such as textual-unit--concept heatmaps, for explaining the detected alignment.

Within this framework, the objective is to define an affinity function that, given a target document and an $SRD$, produces a quantitative alignment measure for each \emph{(textual unit, concept)} pair. Formally, we seek a function $\Lambda$ such that:
\begin{equation}
\Lambda(D,SRD)\in \mathbb{R}^{n\times m},
\label{eq:affinity-matrix}
\end{equation}

\noindent where the component $\Lambda_{ij}(D,SRD)$ quantifies the \emph{affinity} (or alignment) of textual unit $S_i$ in document $D$ with respect to concept $c_j$ defined in the $SRD$. The resulting textual-unit--concept affinity matrix can be interpreted as a projection of the content of $D$ onto the conceptual space induced by $SRD$.

For applications requiring a single score per concept, the matrix $\Lambda(D,SRD)$ can additionally be aggregated along the textual-unit axis to derive a document-level affinity vector, formally defined in Section~\ref{subsec:affinity-levels}.

Overall, this formulation defines the central problem addressed in this work: the construction of a robust and interpretable metric of the conceptual alignment between unstructured documents and a structured reference document.

\subsection{Terms and Reference Term Sets}
\label{subsec:terms-rts}

Affinity computation is based on comparing the content of the target document $D$ with a reference lexicon induced by the $SRD$. To do this, we start with a homogeneous terminological representation for any textual unit.

\paragraph{Term Extraction.}
Let $X$ be a generic textual unit, such that $X \in U(D)\cup U(SRD)$. We denote by $\mathcal{T}(X)$ the \emph{multiset} (set with repeated elements) of terms extracted from $X$ after applying a set of linguistic preprocessing techniques, including orthographic normalization, tokenization, removal of punctuation and \emph{stopwords}, lemmatization, and stemming, among others, as part of the two-phase methodology described in Section \ref{sec:methodology}.
The absolute frequency of a term $t$ in $X$ is defined as:

\begin{equation}
f(t \mid X)=\left|\left\{\,t' \in \mathcal{T}(X)\;:\; t'=t\,\right\}\right|.
\label{eq:term-frequency}
\end{equation}

The \emph{vocabulary} of a multiset of terms is defined as the set of its distinct elements, disregarding multiplicity. For a textual unit $X$, the \emph{term vocabulary} $\mathcal{V}(X)$ is accordingly defined as:

\begin{equation}
	\mathcal{V}(X)=\{\,t \;:\; t \in \mathcal{T}(X)\,\}.
	\label{eq:unique-vocab}
\end{equation}

\noindent The same operator applies directly to any multiset of terms, such as a reference term set $RTS_j$, for which $\mathcal{V}(RTS_j)=\{\,t \;:\; t \in RTS_j\,\}$ follows analogously.

The vocabulary $\mathcal{V}(X)$ is used when only the presence or absence of terms is relevant, without taking into account their frequency of occurrence.

\paragraph{Reference Term Sets by Concept and Terminological-Profile Collection.}

As defined above, the set of textual units in the $SRD$ is represented as $U(SRD)=\{R_1,\dots,R_m\}$, where each unit $R_j$ describes concept $c_j$. The \emph{reference term set} associated with $c_j$ is defined as:

\begin{equation}
	RTS_j = \mathcal{T}(R_j), \qquad j\in\{1,\dots,m\}.
	\label{eq:rts-definition}
\end{equation}

$RTS_j$ is a multiset, not a set, and preserves the term multiplicities of $\mathcal{T}(R_j)$, later required for term frequencies within a concept (e.g., Equation~\ref{eq:term-concept-distribution}). For readability, however, the name $RTS_j$ is retained.

Each $RTS_j$ constitutes the terminological profile of concept $c_j$ and provides the basis for evaluating the affinity of each unit $S_i$ in the target document $D$ with respect to that concept through the function $\Lambda(D,SRD)$.

The \textit{terminological-profile collection}, defined as the set of concept-specific terminological profiles, is denoted as:
\begin{equation}
	\mathcal{TP} = \{RTS_1,\dots,RTS_m\}.
	\label{eq:tp-collection}
\end{equation}

\paragraph{SRD Term Universe.}
The term universe induced by the $SRD$ is defined as the union of the concept-specific vocabularies:

\begin{equation}
\mathcal{V}_{SRD} = \bigcup_{j=1}^{m} \mathcal{V}(R_j)
= \bigcup_{j=1}^{m} \mathcal{V}(RTS_j).
\label{eq:srd-vocab}
\end{equation}

\subsection{Term Typology}
\label{subsec:term-typology}

Document affinity relies on terminological comparison between textual units. However, relevant linguistic evidence may be expressed not only through isolated terms (unigrams) but also through compound terms, including multi-word expressions and bounded co-occurrences. Therefore, this work defines a \emph{term typology} that integrates simple and compound terms uniformly within the multiset $\mathcal{T}(X)$ defined in Section~\ref{subsec:terms-rts}.

\paragraph{Base Token Sequence.}
Let $X \in U(D)\cup U(SRD)$ be a generic textual unit. After linguistic preprocessing, an ordered sequence of normalized \textit{tokens} is obtained:
\begin{equation}
w(X)=( w_1,\dots,w_{L_X}),
\label{eq:base-token-seq}
\end{equation}
where $L_X$ denotes the final number of \textit{tokens} after filtering (removal of punctuation marks, short words, and \emph{stopwords}) and normalization (lemmatization or stemming).

Each \textit{token} $w_k$ is treated as an atomic unit whose specific internal representation is defined in the methodology (Section~\ref{subsubsec:nlp-core}); different term types may draw on different components of that representation, as made explicit at the point of instantiation. Based on ${w}(X)$, the following term types are defined:

\begin{itemize}

\item \textbf{Independent terms (IND):} These correspond to individual \textit{tokens} (unigrams). The multiset of independent terms is defined as:
\begin{equation}
\mathcal{T}^{\mathrm{IND}}(X)=\biguplus_{k=1}^{L_X} \{\, w_k \,\}.
\label{eq:ind-terms}
\end{equation}
where $\uplus$ denotes multiset union.
\
\item \textbf{Bigrams (2G):} These consist of pairs of consecutive \textit{tokens}. The multiset of bigrams is defined as:
\begin{equation}
\mathcal{T}^{\mathrm{2G}}(X)=\biguplus_{k=1}^{L_X-1} \{\, (w_k,w_{k+1}) \,\}.
\label{eq:bigram-terms}
\end{equation}

\item \textbf{Trigrams (3G):} These consist of three consecutive \textit{tokens}. The multiset of trigrams is defined as:
\begin{equation}
\mathcal{T}^{\mathrm{3G}}(X)=\biguplus_{k=1}^{L_X-2} \{\, (w_k,w_{k+1},w_{k+2}) \,\}.
\label{eq:trigram-terms}
\end{equation}

\item \textbf{Co-occurrences (COC):} These are ordered pairs of non-adjacent \textit{tokens} occurring within a bounded window in ${w}(X)$. Unlike bigrams, co-occurrences do not require strict adjacency, but only proximity within the specified window. Furthermore, both directions are considered, such that a pair $(w_k,w_{k+d})$ and its reverse constitute distinct co-occurrences. The multiset of co-occurrences is defined as:

\begin{equation}
\mathcal{T}^{\mathrm{COC}}(X)=
\biguplus_{k=1}^{L_X}\;
\biguplus_{\substack{-N_{sw}-1\le d\le N_{sw}+1\\ d\neq 0,\,+1\\ 1\le k+d\le L_X}}
\left\{\left(w_k,\;w_{k+d}\right)\right\}.
\label{eq:cooc-terms}
\end{equation}

where $d$ denotes the signed displacement between the positions of the two co-occurring \textit{tokens}, and $N_{sw}$ denotes the maximum number of intermediate \textit{tokens} allowed between them, such that $|d|\le N_{sw}+1$. The displacement $d=0$ is excluded by definition because a \textit{token} does not co-occur with itself. Likewise, $d=+1$ (forward adjacency) is explicitly excluded because it is already represented as a bigram (Equation~\ref{eq:bigram-terms}). By contrast, reverse adjacency $d=-1$ is retained as a co-occurrence. The condition $1\le k+d\le L_X$ ensures that all indices remain within the sequence.

\end{itemize}

Based on this typology, the \textit{term multiset of a textual unit} $X$ is defined as the multiset union of the preceding term families:
\begin{equation}
\mathcal{T}(X)=
\mathcal{T}^{\mathrm{IND}}(X)\;\uplus\;
\mathcal{T}^{\mathrm{2G}}(X)\;\uplus\;
\mathcal{T}^{\mathrm{3G}}(X)\;\uplus\;
\mathcal{T}^{\mathrm{COC}}(X),
\label{eq:term-multiset-union}
\end{equation}

The frequency $f(t\mid X)$ (Equation~\ref{eq:term-frequency}) then represents the multiplicity of occurrences of $t$ in the textual unit, regardless of whether $t$ is a simple or compound term.

\subsection{Discriminability, Specificity, Relevance, and Importance}
\label{subsec:disc-spec-rel-imp}

Terms differ in the semantic information they contribute to the characterization of a concept. Terms that occur with similar frequency across multiple concepts, for example, have limited discriminative capacity. By contrast, compound terms ($n$-grams or co-occurrences) generally encode more specific information than isolated terms. To represent these properties, four measures are defined: \emph{discriminability}, \emph{specificity}, \emph{relevance}, and \emph{importance}. These weighting measures are subsequently used to instantiate the affinity function $\Lambda(D,SRD)$.

\paragraph{Discriminability.}
Let $t \in \mathcal{V}_{SRD}$ be a term in the $SRD$ universe (Section~\ref{subsec:terms-rts}). To quantify the distribution of $t$ across the $m$ concepts, we define its \emph{frequency distribution across concepts}, normalized by the total frequency $f$ of $t$ in the $SRD$:
\begin{equation}
p_j(t)=\frac{f(t\mid RTS_j)}{\displaystyle\sum_{k=1}^{m} f(t\mid RTS_k)},
\qquad j\in\{1,\dots,m\}.
\label{eq:term-concept-distribution}
\end{equation}
By construction, $p_j(t)\ge 0$ and $\sum_{j=1}^{m} p_j(t)=1$; therefore, $\{p_j(t)\}_{j=1}^{m}$ constitutes a probability distribution over the concepts. The \emph{discriminability} $\delta(t)\in[0,1]$ quantifies the ability of $t$ to discriminate among concepts and is defined as the complement of the normalized discrete entropy of this distribution:
\begin{equation}
\delta(t)=1-\left(
-\sum_{j=1}^{m} p_j(t)\;
\log_{m}\!\left(p_j(t)\right)
\right),
\label{eq:discriminability}
\end{equation}
using the standard convention $0\cdot\log_m 0 = 0$. The logarithm with base $m$ normalizes entropy to the interval $[0,1]$, which implies that $\delta(t)\in[0,1]$.

Intuitively, $\delta(t)$ approaches 1 when $t$ is predominantly associated with a single concept and approaches 0 when its relative occurrence is similar across all concepts. In the limiting case in which $t$ is associated with only one concept, the distribution $\{p_j(t)\}$ is degenerate, its entropy is zero, and $\delta(t)=1$. In contrast, when the relative frequency of $t$ is identical across all concepts, entropy is maximal and $\delta(t)=0$, indicating zero discriminative power.

\paragraph{Specificity.}
The \emph{specificity} of a term $t$, $\sigma(t\mid RTS_j)\in[0,1]$, measures the semantic information contributed by the term to concept $c_j$ according to the degree of concreteness or restrictiveness of the linguistic pattern it represents. The underlying assumption is that greater structural complexity, and therefore greater contextual information, corresponds to a lower probability of a chance match and, consequently, to greater semantic value. This notion is formalized by associating each term type $\kappa$ with its \emph{pattern-space size}, $a_\kappa(R_j)$. For term types generated combinatorially, this is the number of distinct patterns of that type that can be constructed from the independent terms of the concept. For co-occurrences, it corresponds to the number of patterns of that type actually observed in the descriptor. Therefore, a larger pattern space implies a lower probability of matching a specific pattern by chance and, consequently, a higher assigned specificity.

Let $R_j$ be the reference textual unit describing concept $c_j$, and let $n_j=|\mathcal{V}^{\mathrm{IND}}(R_j)|$ denote the number of distinct independent terms in $R_j$. For each type $\kappa\in K=\{\mathrm{IND},\mathrm{2G},\mathrm{3G},\mathrm{COC}\}$, the pattern-space size $a_\kappa(R_j)$ is defined as:
\begin{equation}
a_{\mathrm{IND}}(R_j)=n_j,\qquad
a_{\mathrm{2G}}(R_j)=\binom{n_j}{2},\qquad
a_{\mathrm{3G}}(R_j)=\binom{n_j}{3},\qquad
a_{\mathrm{COC}}(R_j)=\big|\mathcal{V}^{\mathrm{COC}}(R_j)\big|.
\label{eq:specificity-potentials}
\end{equation}

\noindent Independent terms contribute $n_j$ patterns, corresponding to the \textit{tokens} themselves. Bigrams and trigrams contribute the numbers of pairs and triples of independent terms that can be formed, namely $\binom{n_j}{2}$ and $\binom{n_j}{3}$, respectively. For co-occurrences, $a_{\mathrm{COC}}(R_j)$ denotes the number of distinct co-occurrences actually observed in $R_j$ (Equation~\ref{eq:cooc-terms}), that is, the cardinality of its co-occurrence vocabulary $\mathcal{V}^{\mathrm{COC}}(R_j)$. Unlike the pattern spaces of bigrams and trigrams, which are fully determined by $n_j$, the co-occurrence pattern space also depends on the proximity window $N_{sw}$ and on the specific arrangement of terms in the text. It is therefore quantified directly from the observed co-occurrences, which represent the pattern space effectively available in each descriptor.

The specificity of a term $t$ of type $\kappa(t)$ is then defined as the fraction of the pattern space corresponding to its type, normalized across all term types:
\begin{equation}
\sigma(t\mid RTS_j) = \sigma_{\kappa(t)}(R_j),
\qquad
\sigma_\kappa(R_j) = \frac{a_\kappa(R_j)}{\displaystyle\sum_{\kappa' \in K} a_{\kappa'}(R_j)}.
\label{eq:specificity}
\end{equation}

\noindent This definition makes specificity concept-dependent through $n_j$. By definition, the following normalization properties are satisfied:
\begin{equation}
\sigma_\kappa(R_j)\in[0,1],\qquad
\sum_{\kappa\in K}\sigma_\kappa(R_j)=1,
\label{eq:specificity-normalization}
\end{equation}

\noindent ensuring that specificity is bounded within $[0,1]$ and distributed across term types in proportion to the relative size of their pattern spaces.

\paragraph{Relevance.}

For any $t \in \mathcal{V}_{SRD}$, define its \emph{relevance} to concept $c_j$ as:

\begin{equation}
\eta(t\mid RTS_j)=
\begin{cases}
0, & t \notin \mathcal{V}(RTS_j),\\
\sigma(t\mid RTS_j), & t \in \mathcal{V}(RTS_j)
\end{cases}
\label{eq:relevance}
\end{equation}
In effect, $\eta(t\mid RTS_j)$ functions as a membership filter for the concept lexicon, weighting the contribution of the term according to the specificity associated with its typology.

\paragraph{Importance.}
The \emph{importance} $\varphi(t\mid RTS_j)\in[0,1]$ combines intra-concept relevance, represented by filtered specificity, with inter-concept discriminability:
\begin{equation}
\varphi(t\mid RTS_j)=\eta(t\mid RTS_j)\cdot \delta(t).
\label{eq:importance}
\end{equation}
A term therefore has high importance for a concept when it belongs to the concept’s terminological profile (non-zero relevance and high specificity), and it discriminates that concept from the remaining concepts (high discriminability). Since $\eta(t\mid RTS_j)\in[0,1]$ (because $\sigma$ lies within this interval, Equation~\ref{eq:specificity-normalization}) and $\delta(t)\in[0,1]$, their product satisfies $\varphi(t\mid RTS_j)\in[0,1]$. These measures are subsequently used to define the weight of matching terms and, consequently, the affinity $\Lambda_{ij}(D,SRD)$.

\subsection{Affinity at the Term, Textual-Unit, and Document Levels}
\label{subsec:affinity-levels}

Based on the terminological profiles $RTS_j$ (Section~\ref{subsec:terms-rts}), the measures defined in Section~\ref{subsec:disc-spec-rel-imp} and combined through the importance function $\varphi(\cdot)$, and the terms extracted from a target document $D$, affinity is defined at three levels: term affinity, textual-unit affinity, and aggregated document-level affinity. In each case, affinity is interpreted as a nonnegative measure of alignment obtained by aggregating terminological matches between $D$ and the $SRD$, weighted by frequency, specificity, and discriminability.

\paragraph{Elementary affinity contribution.}
Let $S_i \in U(D)$ be a textual unit in the target document, and let $t \in \mathcal{V}(S_i)$ be a term contained in that unit. The contribution, or elementary affinity, of term $t$ to concept $c_j$ is defined as the product of its frequency in $S_i$ (Equation~\ref{eq:term-frequency}) and its importance with respect to the concept (Equation~\ref{eq:importance}):
\begin{equation}
\ell(t;S_i,RTS_j) \;=\; f(t\mid S_i)\cdot \varphi(t\mid RTS_j).
\label{eq:term-level-affinity}
\end{equation}
If $t \notin \mathcal{V}(RTS_j)$, then $\eta(t\mid RTS_j)=0$ and, consequently, $\varphi(t\mid RTS_j)=0$, so the term does not contribute to alignment with $c_j$.

\paragraph{Textual-Unit-Level Affinity.}
The affinity of a textual unit $S_i$ with respect to concept $c_j$ is defined by aggregating the contributions of its terms. To obtain a measure that is comparable across units of different lengths, the formulation normalizes by the total number of terms in the unit, including multiplicity, $|\mathcal{T}(S_i)|$:
\begin{equation}
\Lambda_{ij}(D,SRD)
\;=\;
\frac{1}{|\mathcal{T}(S_i)|}
\sum_{t \in \mathcal{V}(S_i)}
\ell(t;S_i,RTS_j)
\;=\;
\frac{1}{|\mathcal{T}(S_i)|}
\sum_{t \in \mathcal{V}(S_i)}
f(t\mid S_i)\cdot \varphi(t\mid RTS_j).
\label{eq:unit-level-affinity}
\end{equation}
Hence, each component $\Lambda_{ij}(D,SRD)$ of the affinity matrix (Equation~\ref{eq:affinity-matrix}) can be interpreted as the \emph{mean importance}, weighted by frequency, of the terms in $S_i$ that are relevant to concept $c_j$. Since $\varphi(t\mid RTS_j)\in[0,1]$, the preceding normalization yields $\Lambda_{ij}(D,SRD)\in[0,1]$.

\paragraph{Aggregated Document-Level Affinity.}
The affinity of document $D$ with respect to each concept $c_j$ is obtained by aggregating the affinities of its textual units. Consistently with Section~\ref{subsec:problem-statement}, we define:
\begin{equation}
\lambda_j(D,SRD)=\mathcal{A}\big(\Lambda_{1j}(D,SRD),\dots,\Lambda_{nj}(D,SRD)\big),
\label{eq:doc-level-affinity}
\end{equation}
The operator $\mathcal{A}(\cdot)$ is selected according to the notion of alignment adopted: sustained alignment throughout the document (mean), strong presence (maximum or high percentile), or accumulated evidence (sum).

The document affinity vector is therefore composed of the aggregated affinities for each concept:
\begin{equation}
\lambda(D,SRD)=\left[\lambda_1(D,SRD),\dots,\lambda_m(D,SRD)\right].
\label{eq:doc-affinity-vector}
\end{equation}

\subsection{Absolute Affinity, Reference Affinity, and Relative Affinity}
\label{subsec:absolute-ref-relative}

The definitions introduced above yield both the textual-unit-level affinity matrix $\Lambda(D,SRD)\in\mathbb{R}^{n\times m}$ and the document-level affinity vector $\lambda(D,SRD)\in\mathbb{R}^{m}$. However, the magnitude of these affinity values depends on the $SRD$ itself, including its terminological richness and the distribution of terms across concepts, as well as on the aggregation operator $\mathcal{A}(\cdot)$. Therefore, no standard reference values are available for direct comparison, which limits the interpretability of the resulting scores. To provide a more consistent basis for interpretation and comparison, three types of affinity are distinguished: \emph{absolute affinity}, \emph{reference affinity}, and \emph{relative affinity}.

\paragraph{Absolute Affinity.}
The term \emph{absolute affinity} denotes the values produced directly by the model, namely each component $\Lambda_{ij}(D,SRD)$ (Equation~\ref{eq:unit-level-affinity}) and the aggregated affinity per concept $\lambda_j(D,SRD)$ (Equation~\ref{eq:doc-level-affinity}). In particular, $\Lambda_{ij}(D,SRD)\in[0,1]$ by construction, whereas the range of $\lambda_j(D,SRD)$ depends on the aggregation operator $\mathcal{A}(\cdot)$. For example, if $\mathcal{A}$ is the sum, $\lambda_j$ increases with $n$; if $\mathcal{A}$ is the mean, $\lambda_j\in[0,1]$.

\paragraph{Reference Affinity.}
A concept-specific \emph{reference affinity} is defined to provide an interpretive anchor. For each concept $c_j$, its descriptive text $R_j$ in the $SRD$ is considered the reference document and is treated as a document containing a single textual unit:

\begin{equation}
	D^{\mathrm{ref}}_j \;:\; U(D^{\mathrm{ref}}_j)=\{R_j\}.
	\label{eq:ref-document}
\end{equation}

The reference affinity of concept $c_j$ is then defined as:

\begin{equation}
	\lambda^{\mathrm{ref}}_j(SRD) = \lambda_j(D^{\mathrm{ref}}_j,SRD),
	\label{eq:ref-affinity}
\end{equation}

\noindent which, because the reference document contains a single textual unit, is equivalent to the corresponding textual-unit--concept affinity:
\begin{equation}
	\lambda^{\mathrm{ref}}_j(SRD)
	\;=\;
	\Lambda_{1j}(D^{\mathrm{ref}}_j,SRD)
	\;=\;
	\frac{1}{|\mathcal{T}(R_j)|}
	\sum_{t\in \mathcal{V}(R_j)} f(t\mid R_j)\cdot \varphi(t\mid RTS_j).
	\label{eq:ref-affinity-expanded}
\end{equation}

The reference affinity $\lambda^{\mathrm{ref}}_j(SRD)$ thus represents the intrinsic alignment level of the concept descriptor with its terminological profile $RTS_j$ and thereby provides a concept-specific reference scale for each $c_j$.

This reference affinity is strictly positive under a mild regularity condition on the $SRD$: each concept $c_j$ contains at least one term with nonzero discriminability, that is, $\delta(t) > 0$ for some $t \in \mathcal{V}(R_j)$. Since $\eta(t \mid RTS_j) = \sigma(t \mid RTS_j) > 0$ for every term $t \in \mathcal{V}(R_j)$ actually observed in the descriptor (Section~\ref{subsec:disc-spec-rel-imp}), this condition alone suffices to guarantee $\lambda^{\mathrm{ref}}_j(SRD) > 0$. The condition is expected to hold whenever the $SRD$ is informative enough to distinguish its concepts: a descriptor composed exclusively of terms with uniform frequency across all concepts would contribute no discriminative evidence to any concept, including its own. In the degenerate case where this condition fails for some $c_j$, the relative affinity $\lambda^{\mathrm{rel}}_j(D,SRD)$ (Equation~\ref{eq:relative-affinity}) is left undefined for that concept, while the corresponding absolute affinity $\lambda_j(D,SRD)$ remains well defined and can be reported in its place.

\paragraph{Relative Affinity.}
Based on the reference affinity, the \emph{relative affinity} of the target document $D$ with respect to concept $c_j$ is defined, whenever $\lambda^{\mathrm{ref}}_j(SRD) > 0$, as the normalization ratio:

\begin{equation}
	\lambda^{\mathrm{rel}}_j(D,SRD)
	\;=\;
	\frac{\lambda_j(D,SRD)}{\lambda^{\mathrm{ref}}_j(SRD)},
	\qquad j\in\{1,\dots,m\}.
	\label{eq:relative-affinity}
\end{equation}
Similarly, the relative affinity vector is defined as:
\begin{equation}
	\lambda^{\mathrm{rel}}(D,SRD)
	=
	\left[\lambda^{\mathrm{rel}}_1(D,SRD),\dots,\lambda^{\mathrm{rel}}_m(D,SRD)\right].
	\label{eq:relative-affinity-vector}
\end{equation}

A value of $\lambda^{\mathrm{rel}}_j(D,SRD)$ close to $1$ indicates an alignment level comparable to that of the reference text $R_j$. Lower values indicate weaker alignment, whereas values greater than $1$ may occur when the target document accumulates repeated or strong semantic evidence associated with concept $c_j$, depending on the aggregation operator $\mathcal{A}$.

\subsection{Model Output and Intermediate Elements}
\label{subsec:model-output-elements}

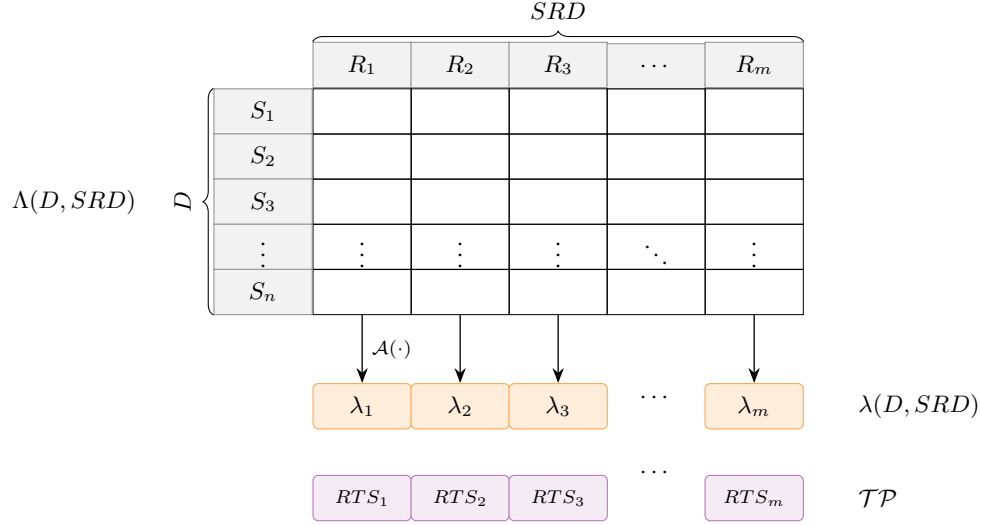
\begin{figure}[t]
	\centering
	\begin{tikzpicture}[
	font=\small,
	>={Stealth[length=2mm]},
	cell/.style={draw, minimum width=13mm, minimum height=6mm, inner sep=1pt, line width=0.3pt},
	hcell/.style={cell, fill=black!5, draw=black!45},
	outbox/.style={draw, rounded corners=2pt, fill=orange!14, draw=orange!70,
		minimum width=13mm, minimum height=6mm},
	csbox/.style={draw, rounded corners=2pt, fill=violet!8, draw=violet!55,
		minimum width=13mm, minimum height=6mm, font=\scriptsize},
	arr/.style={->, line width=0.5pt},
	bracelbl/.style={font=\small\bfseries}
	]
	
	\matrix (grid) [matrix of nodes,
	nodes in empty cells,
	nodes={cell},
	column sep=-\pgflinewidth, row sep=-\pgflinewidth,
	row 1/.style={nodes={hcell}}
	]
	{
		$R_1$ & $R_2$ & $R_3$ & $\cdots$ & $R_m$ \\
		&       &       &         &       \\
		&       &       &         &       \\
		&       &       &         &       \\
		$\vdots$ & $\vdots$ & $\vdots$ & $\ddots$ & $\vdots$ \\
		&       &       &         &       \\
	};
	
	\node[hcell, left=0mm of grid-2-1] (r1) {$S_1$};
	\node[hcell, left=0mm of grid-3-1] (r2) {$S_2$};
	\node[hcell, left=0mm of grid-4-1] (r3) {$S_3$};
	\node[hcell, left=0mm of grid-5-1] (rd) {$\vdots$};
	\node[hcell, left=0mm of grid-6-1] (rn) {$S_n$};
	
	\draw[decorate, decoration={brace, amplitude=4pt}]
	(grid-1-1.north west) -- (grid-1-5.north east)
	node[midway, above=5pt, bracelbl] {$SRD$};
	
	\draw[decorate, decoration={brace, amplitude=4pt, mirror}]
	(r1.north west) -- (rn.south west)
	node[midway, left=5pt, bracelbl, rotate=90, anchor=south] {$D$};
	
	\node[font=\small, anchor=east] at ($(r1.west)!0.5!(rn.west)+(-9mm,0)$) {$\Lambda(D,SRD)$};
	
	\node[outbox, below=9mm of grid-6-1] (l1) {$\lambda_1$};
	\node[outbox, below=9mm of grid-6-2] (l2) {$\lambda_2$};
	\node[outbox, below=9mm of grid-6-3] (l3) {$\lambda_3$};
	\node[font=\small, below=9mm of grid-6-4] (ld) {$\cdots$};
	\node[outbox, below=9mm of grid-6-5] (lm) {$\lambda_m$};
	
	\draw[arr] (grid-6-1.south) -- (l1.north)
	node[midway, right, font=\scriptsize] {$\mathcal{A}(\cdot)$};
	\draw[arr] (grid-6-2.south) -- (l2.north);
	\draw[arr] (grid-6-3.south) -- (l3.north);
	\draw[arr] (grid-6-5.south) -- (lm.north);
	
	\node[font=\small, anchor=west] at ([xshift=6mm]lm.east) {$\lambda(D,SRD)$};
	
	\node[csbox, below=6mm of l1] (c1) {$RTS_1$};
	\node[csbox, below=6mm of l2] (c2) {$RTS_2$};
	\node[csbox, below=6mm of l3] (c3) {$RTS_3$};
	\node[font=\small, below=6mm of ld] (cd) {$\cdots$};
	\node[csbox, below=6mm of lm] (cm) {$RTS_m$};
	
	\node[font=\small, anchor=west] at ([xshift=6mm]cm.east) {$\mathcal{TP}$};
	
	\end{tikzpicture}
	\caption{Schematic representation of the textual-unit--concept affinity matrix $\Lambda(D,SRD)$. Rows correspond to the textual units $S_i \in U(D)$ and columns to the concepts $c_j \in SRD$, each described by its reference textual unit $R_j$ and characterized by its terminological profile $RTS_j$. The aggregation operator $\mathcal{A}(\cdot)$ reduces each column of $\Lambda(D,SRD)$ to the document-level affinity $\lambda_j(D,SRD)$, jointly forming the vector $\lambda(D,SRD)$.}
	\label{fig:affinity-matrix}
	
\end{figure}

The preceding sections induce, from the $SRD$, a reference model $M$ that associates each concept $c_j$ with its terminological profile $RTS_j$ and the weighting measures introduced in Section~\ref{subsec:disc-spec-rel-imp}, together with the concept-specific reference affinity $\lambda^{\mathrm{ref}}_j(SRD)$. Thus, given a target document $D$, this model determines the textual-unit--concept affinity matrix $\Lambda(D,SRD)$ and its aggregated absolute and relative vectors, $\lambda(D,SRD)$ and $\lambda^{\mathrm{rel}}(D,SRD)$, as illustrated in Figure~\ref{fig:affinity-matrix}. These elements constitute the formal output of the model and provide a quantitative and interpretable representation of the conceptual alignment of document $D$ with respect to the $SRD$. The complete notation is collected in Appendix~\ref{sec:glossary}.


\section{Method}\label{sec:methodology}

The methodology underlying ITL implements the affinity quantification formalized in Section~\ref{sec:problem} through a reproducible procedure. Its design satisfies three requirements: 1) domain and language independence, allowing the $SRD$ to be instantiated from any concept-structured reference framework regardless of the language in which it is written; 2) interpretability, by preserving traceability from each affinity value to the terminological evidence on which it is based; and 3) applicability to unstructured documents, such as PDFs, which are common in the intended application scenarios.

ITL is structured in two complementary phases: a \emph{reference-model induction phase}, in which the reference model is induced from the $SRD$, and a \emph{document-alignment phase}, in which the affinity of a target document is computed.

During the \emph{induction phase} (Section~\ref{subsec:phase1}), a \emph{reference model} is generated from the $SRD$. This model is a structured and attributed representation that associates each concept $c_j$ with its terminological profile $RTS_j$ and the weighting measures (discriminability, specificity, relevance, and importance) defined in Section~\ref{subsec:disc-spec-rel-imp}. The model is computed once for each $SRD$ and subsequently reused to evaluate any target document.

During the \emph{alignment phase} (Section~\ref{subsec:phase2}), a target document $D$ is segmented into textual units and processed using the same linguistic-processing procedure employed during induction. Its terms are then compared with the reference model to compute the affinity matrix $\Lambda(D,SRD)$ and the corresponding document-level aggregations (Equations~\ref{eq:unit-level-affinity}, \ref{eq:doc-level-affinity}, and \ref{eq:relative-affinity}).

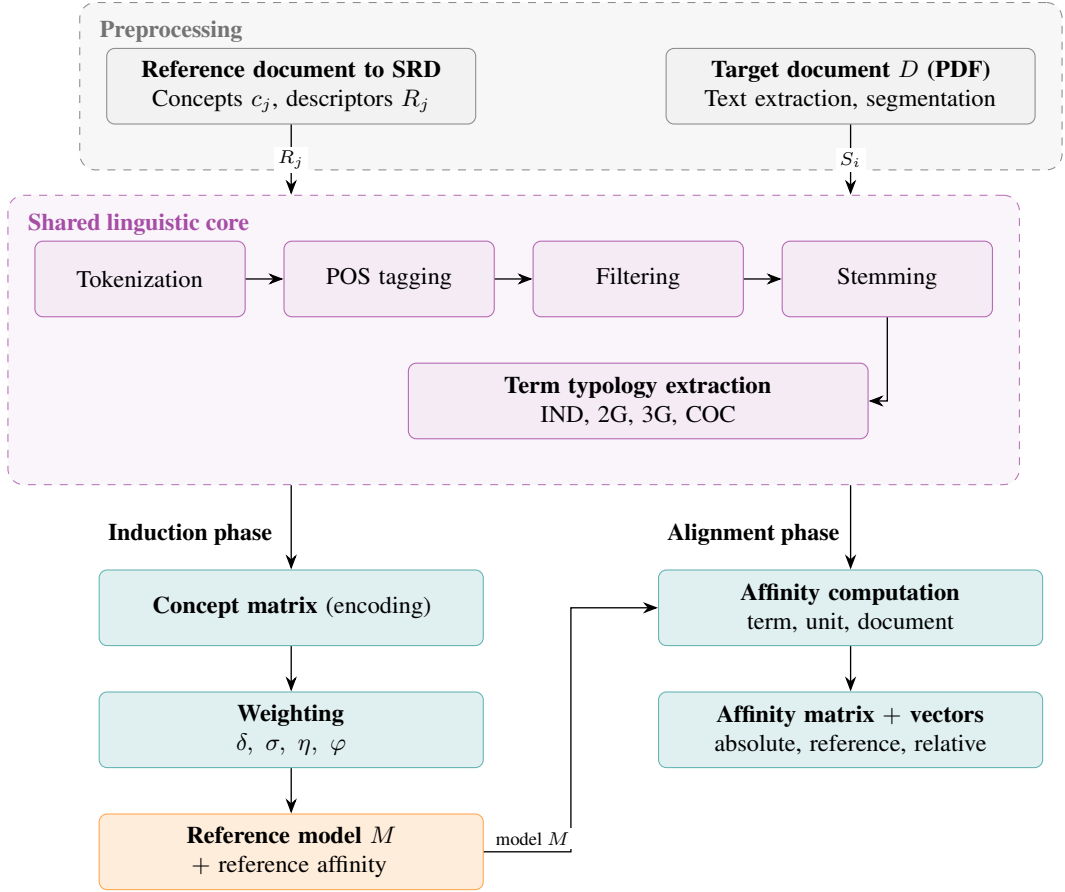
\begin{figure}[t]
\centering
\begin{tikzpicture}[
  font=\small,
  >={Stealth[length=2mm]},
  box/.style={draw, rounded corners=3pt, align=center, inner sep=4pt,
              minimum height=10mm, line width=0.4pt},
  input/.style={box, fill=black!5, draw=black!45, text width=46mm},
  core/.style={box, fill=violet!8, draw=violet!60, text width=25mm},
  train/.style={box, fill=teal!12, draw=teal!60, text width=48mm},
  model/.style={box, fill=orange!14, draw=orange!70, text width=48mm},
  phaselbl/.style={font=\small\bfseries},
  edgelbl/.style={font=\scriptsize, inner sep=1.5pt, fill=white},
  arr/.style={->, line width=0.5pt}
]

\def\LX{0}      
\def\RX{7.4}    

\node[input] (srd) at (\LX,0) {\textbf{Reference document to SRD}\\[1pt]\footnotesize Concepts $c_j$, descriptors $R_j$};
\node[input] (doc) at (\RX,0) {\textbf{Target document $D$ (PDF)}\\[1pt]\footnotesize Text extraction, segmentation};

\node[core] (tok) at (\LX-2.0,-2.55) {Tokenization};
\node[core, right=5mm of tok] (pos) {POS tagging};
\node[core, right=5mm of pos] (fil) {Filtering};
\node[core, right=5mm of fil] (ste) {Stemming};
\node[core, text width=58mm, below=6mm of fil.south, anchor=north] (typ)
   {\textbf{Term typology extraction}\\[1pt]\footnotesize IND, 2G, 3G, COC};

\node[train] (enc) at (\LX,-6.9) {\textbf{Concept matrix} (encoding)};
\node[train, below=6mm of enc] (wgt) {\textbf{Weighting}\\[1pt]\footnotesize $\delta,\ \sigma,\ \eta,\ \varphi$};
\node[model, below=6mm of wgt] (mod) {\textbf{Reference model $M$}\\[1pt]\footnotesize $+$ reference affinity};

\node[train] (aff) at (\RX,-6.9) {\textbf{Affinity computation}\\[1pt]\footnotesize term, unit, document};
\node[train, below=6mm of aff] (mat) {\textbf{Affinity matrix $+$ vectors}\\[1pt]\footnotesize absolute, reference, relative};

\node[phaselbl, anchor=south west] at (enc.north west |- enc.north) [yshift=2mm] {Induction phase};
\node[phaselbl, anchor=south west] at (aff.north west |- aff.north) [yshift=2mm] {Alignment phase};

\begin{scope}[on background layer]
  \node[draw, dashed, rounded corners=6pt, line width=0.4pt, fill=black!3, draw=black!45,
        inner ysep=6mm, inner xsep=3.5mm, fit=(srd)(doc)] (preproc) {};
  \node[draw, dashed, rounded corners=6pt, line width=0.4pt, fill=violet!4, draw=violet!55,
        inner ysep=6mm, inner xsep=3.5mm, fit=(tok)(ste)(typ)] (coreb) {};
\end{scope}
\node[phaselbl, black!55, anchor=north west] at (preproc.north west) [xshift=1.5mm, yshift=-1.2mm] {Preprocessing};
\node[phaselbl, violet!70, anchor=north west] at (coreb.north west) [xshift=1.5mm, yshift=-1.2mm] {Shared linguistic core};

\draw[arr] (srd.south) -- node[edgelbl] {$R_j$} (srd.south |- coreb.north);
\draw[arr] (doc.south) -- node[edgelbl] {$S_i$} (doc.south |- coreb.north);

\draw[arr] (tok) -- (pos);
\draw[arr] (pos) -- (fil);
\draw[arr] (fil) -- (ste);
\draw[arr] (ste.south) |- (typ.east);

\draw[arr] (coreb.south -| enc.north) -- (enc.north);
\draw[arr] (coreb.south -| aff.north) -- (aff.north);

\draw[arr] (enc) -- (wgt);
\draw[arr] (wgt) -- (mod);

\draw[arr] (aff) -- (mat);

\draw[arr] (mod.east) -| ($(mod.east)!0.5!(aff.west)+(0,0)$)
      node[edgelbl, pos=0.28, above, yshift=0.3mm] {model $M$} |- (aff.west);

\end{tikzpicture}
\caption{ITL Architecture.}
\label{fig:architecture}
\end{figure}

Both phases share a common linguistic-processing core, which consists of tokenization, tagging, filtering, \emph{stemming}, and term-typology extraction. The principal distinction lies in the processing unit: during induction, the unit is the descriptor $R_j$ associated with each concept, whereas during alignment it is each textual unit $S_i$ of the target document. Figure~\ref{fig:architecture} summarizes this architecture and the ITL processing flow.

Section~\ref{subsec:srd-construction} describes the preprocessing procedure shared by both phases. Sections~\ref{subsec:phase1} and~\ref{subsec:phase2} detail the induction and alignment phases, respectively.

\subsection{Preprocessing and Construction of the \texorpdfstring{$SRD$}{SRD}}
\label{subsec:srd-construction}

The starting point is a reference document that must be structured into the concept set $\mathcal{C}=\{c_1,\dots,c_m\}$ by associating each concept $c_j$ with a descriptive textual unit $R_j$ (Section~\ref{subsec:problem-statement}). The result of this preprocessing stage is the $SRD$, represented as $U(SRD)=\{R_1,\dots,R_m\}$, which constitutes the input to the induction phase. The specific segmentation of the reference document into concepts and descriptors depends on its internal structure and is detailed for the SDG case in Section~\ref{sec:case}.

Similarly, during the alignment phase, the target document $D$ is transformed into the sequence of textual units $U(D)=\{S_1,\dots,S_n\}$ through sentence segmentation. When $D$ is provided in an unstructured format, this segmentation is preceded by a text-extraction stage.

\subsection{Reference-Model Induction Phase: Modeling the \texorpdfstring{$SRD$}{SRD}}
\label{subsec:phase1}

The induction phase takes the $SRD$ as input and produces the reference model $M$. The procedure is summarized in Algorithm~\ref{alg:training} and described in detail below.

\begin{algorithm}[t]
\caption{Reference-model induction phase: modeling the $SRD$}
\label{alg:training}
\DontPrintSemicolon
\KwIn{$SRD=\{R_1,\dots,R_m\}$: structured reference document with $m$ concept descriptors; 
\texttt{min\_short\_len}: maximum discarded token length; $lsw$: \emph{stopword} list; \texttt{n\_sw\_coo}: co-occurrence window.}

\KwOut{$M$: reference model (terms, types, importance per concept, and concept-specific reference affinities).}
\BlankLine
\ForEach{descriptor $R_j \in SRD$}{
  $T \leftarrow \mathrm{Tokenization}(R_j)$ \tcp*{tokens}
  $T_{g} \leftarrow \mathrm{Tagging}(T)$ \tcp*{POS-tagged tokens}
  $T_{f} \leftarrow \mathrm{Filtering}(T_{g}, lsw, \texttt{min\_short\_len})$ \tcp*{filtered tokens}
  $T_{s} \leftarrow \mathrm{Stemming}(T_{f})$ \tcp*{IND term $=$ (stem, tag)}
  $\mathcal{T}(R_j) \leftarrow T_{s} \uplus \mathrm{Extract\_2G\_3G\_COC}(T_{s}, \texttt{n\_sw\_coo})$ \tcp*{term typology}
}
$\mathbf{F} \leftarrow \mathrm{Encoding}\big(\{\mathcal{T}(R_j)\}_{j=1}^{m}\big)$ \tcp*{concept matrix (Eq.~\ref{eq:concept-matrix})}
$M \leftarrow \mathrm{Weighting}(\mathbf{F})$ \tcp*{$\delta,\sigma,\eta,\varphi$ (Eq.~\ref{eq:discriminability},~\ref{eq:specificity},~\ref{eq:relevance},~\ref{eq:importance})}
\Return $M$\;
\end{algorithm}

\subsubsection{Linguistic-Processing Core}
\label{subsubsec:nlp-core}

Each descriptor $R_j$ is processed through the following pipeline, whose objective is to transform free text into the normalized sequence of \textit{tokens} $w(R_j)$ (Equation~\ref{eq:base-token-seq}) used for term extraction.

\paragraph{Tokenization.}
The text associated with each concept is divided into elementary \textit{tokens}, including words, punctuation marks, numbers, and symbols, thereby producing a token sequence for each concept.

\paragraph{Part-of-speech tagging (\emph{POS tagging}).}
Each \textit{token} is assigned a grammatical tag (\emph{part-of-speech}) identifying its category, such as noun, verb, adjective, or adverb. This tag serves two purposes: it provides a lexical-selection criterion during filtering and, for independent terms, forms part of the identity of the term itself (see the \emph{stemming} step below). The specific tagging system and tag inventory adopted in the reference instantiation are specified in Section~\ref{sec:case}.

\paragraph{Filtering.}
The token sequence is filtered to retain only those elements with semantic information relevant to the characterization of concepts. The filtering procedure applies four criteria:
\begin{enumerate}
\item \emph{Removal of punctuation marks, numbers, and symbols.} \textit{Tokens} that are not words are removed.
\item \emph{Removal of stopwords.} Words contained in the predefined list $lsw$, including articles, prepositions, conjunctions, pronouns, and other words without specific semantic content, are removed.
\item \emph{Removal of short words.} \textit{Tokens} whose length is less than or equal to the parameter \texttt{min\_short\_len} are removed. This criterion is applied before \emph{stemming}, using the surface form of each word.
\item \emph{Part-of-speech selection.} Only \textit{tokens} whose grammatical categories transmit conceptual content (nouns, verbs, adjectives, and adverbs) are retained.
\end{enumerate}
All retained words are additionally normalized to lowercase.

\paragraph{\emph{Stemming}.}
A \emph{stemming} algorithm is applied to the filtered \textit{tokens} to reduce each word to its stem. In the methodological instantiation, each \textit{token} $w_k$ in the base sequence $w(R_j)$ (Equation~\ref{eq:base-token-seq}) is represented by the pair $(\ell_k, g_k)$, where $\ell_k$ denotes the stem generated by the \emph{stemmer} and $g_k$ denotes the part-of-speech tag assigned in the preceding step.

Term identity depends on term type. For independent terms, the complete pair $(\ell_k, g_k)$ defines the term, so the same stem associated with different grammatical tags, for example as a noun and as a verb, produces distinct independent terms with distinct weighting measures. This distinction is particularly relevant for single-word terms, for which the same lexical form can take on different meanings depending on its syntactic function.

For each concept, the output is therefore an ordered sequence of independent terms that preserves their order of occurrence in the source text. This ordering is required because compound terms are derived from contiguous or nearby combinations of these independent terms.

\paragraph{Term-Typology Extraction.}
The sequence of independent terms is used to generate the four term families defined in Section~\ref{subsec:term-typology}: independent terms (\textbf{IND}), bigrams (\textbf{2G}), trigrams (\textbf{3G}), and co-occurrences (\textbf{COC}). Bigrams and trigrams correspond to subsequences of two and three consecutive independent terms, respectively. Co-occurrences correspond to pairs of non-consecutive independent terms separated by a bounded number of positions determined by the parameter \texttt{n\_sw\_coo}, which corresponds to $N_{sw}$ in Equation~\ref{eq:cooc-terms}. The associated window is bidirectional, generates ordered pairs, and excludes only direct forward adjacency ($d{=}{+}1$), which is already represented by bigrams. Reverse adjacency ($d{=}{-}1$) is retained as a co-occurrence.

Compound terms (2G, 3G, and COC) differ from independent terms in that they are formed by concatenating only the stems $\ell_k$, without incorporating the part-of-speech tag. Accordingly, Equations~\ref{eq:bigram-terms}, \ref{eq:trigram-terms}, and~\ref{eq:cooc-terms} are instantiated by substituting each token $w_k$ with its stem $\ell_k$, whereas Equation~\ref{eq:ind-terms} is instantiated with the full pair $w_k=(\ell_k,g_k)$. This introduces an asymmetry with respect to independent terms: the same stem associated with different grammatical tags produces distinct independent terms but the same bigram, trigram, or co-occurrence. This design prevents fragmentation of the same multi-word pattern across different combinations of constituent tags, which would otherwise increase model sparsity.

\subsubsection{Concept Matrix (\emph{Encoding})}
\label{subsubsec:encoding}

Once the terms associated with all concepts have been extracted, the frequencies $f(t\mid RTS_j)$ already introduced in Section~\ref{subsec:disc-spec-rel-imp} are assembled into a single \emph{concept matrix} for computational convenience. Formally, given the term universe $\mathcal{V}_{SRD}$ (Equation~\ref{eq:srd-vocab}), the concept matrix $\mathbf{F}\in\mathbb{N}^{|\mathcal{V}_{SRD}|\times m}$ has components:
\begin{equation}
	F_{tj}=f(t\mid RTS_j),\qquad t\in\mathcal{V}_{SRD},\ j\in\{1,\dots,m\},
	\label{eq:concept-matrix}
\end{equation}
where each row corresponds to a term, with its type $\kappa(t)$, and each column corresponds to a concept $c_j$. This tabular representation is used in the remainder of the pipeline to compute the weighting measures efficiently across the full vocabulary.

\subsubsection{Term Weighting and Model Induction}
\label{subsubsec:weighting}

From the concept matrix, the measures that determine the contribution of each term to alignment are computed for every (term, concept) pair: \emph{discriminability} $\delta(t)$ (Equation~\ref{eq:discriminability}), \emph{specificity} $\sigma(t\mid RTS_j)$ (Equation~\ref{eq:specificity}), \emph{relevance} $\eta(t\mid RTS_j)$ (Equation~\ref{eq:relevance}), and \emph{importance} $\varphi(t\mid RTS_j)$ (Equation~\ref{eq:importance}). Their formal definitions are provided in Section~\ref{subsec:disc-spec-rel-imp}.

Discriminability is computed from the normalized entropy of the distribution of a term across the $m$ concepts (Equation~\ref{eq:discriminability}). Specificity, in contrast, is computed according to term type and depends on the concept (Equations~\ref{eq:specificity-potentials}--\ref{eq:specificity}).

The resulting reference model $M$ is represented as a table that associates each term with its type $\kappa(t)$ and, for each concept $c_j$, its importance $\varphi(t\mid RTS_j)$, together with the concept-specific reference affinity $\lambda^{\mathrm{ref}}_j(SRD)$ (Equation~\ref{eq:ref-affinity}), subsequently used as an interpretive anchor during the alignment phase.

\subsection{Document-Alignment Phase: Affinity Computation}
\label{subsec:phase2}

Given a target document $D$ and the reference model $M$, the alignment phase quantifies the affinity between $D$ and the $SRD$. The document is segmented into textual units $U(D)=\{S_1,\dots,S_n\}$, and each unit is processed using the same linguistic-processing core employed during the induction phase (Section~\ref{subsubsec:nlp-core}). The only distinction is that the processing unit is the sentence $S_i$ rather than the descriptor $R_j$. This procedure produces, for each textual unit, the multiset of terms $\mathcal{T}(S_i)$ and their corresponding frequencies.

The extracted terms and the reference model are then used to compute affinity at the three levels defined in Section~\ref{subsec:affinity-levels}: term level (Equation~\ref{eq:term-level-affinity}), textual-unit level (Equation~\ref{eq:unit-level-affinity}), and document level (Equations~\ref{eq:doc-level-affinity}--\ref{eq:doc-affinity-vector}). Relative affinity is subsequently obtained according to Equation~\ref{eq:relative-affinity}. The complete procedure is summarized in Algorithm~\ref{alg:application}.

At the textual-unit level, affinity is computed by weighting each matching term according to its frequency in the unit, $f(t\mid S_i)$, and its importance with respect to the concept, $\varphi(t\mid RTS_j)$, and then normalizing by the size of the unit (Equation~\ref{eq:unit-level-affinity}). Consequently, $\Lambda_{ij}(D,SRD)\in[0,1]$.

At the document level, affinity is obtained by aggregating unit-level affinities through the operator $\mathcal{A}(\cdot)$ (Equation~\ref{eq:doc-level-affinity}). This operator is a configurable parameter of the method and is selected according to the notion of alignment being represented: the mean characterizes sustained alignment throughout the document; the maximum or a high percentile represents strong localized presence; and the sum represents accumulated evidence.

Relative affinity (Equation~\ref{eq:relative-affinity}) normalizes the aggregated affinity by the concept-specific reference affinity $\lambda^{\mathrm{ref}}_j(SRD)$. Thus, a value close to $1$ indicates an alignment level comparable to that of the concept descriptor itself. This interpretation is preserved when $\mathcal{A}(\cdot)$ is scale-invariant with respect to the number of textual units, as in the case of the mean; the maximum and high percentiles remain bounded in $[0,1]$ but tend to increase with $n$, and are therefore only approximately scale-invariant. By contrast, when an extensive operator such as the sum is used, aggregated affinity increases with document length and relative affinity is not upper-bounded. In this case, the sum represents accumulated absolute evidence rather than bounded relative affinity.

\begin{algorithm}[t]
\caption{Document-alignment phase: computing the affinity of $D$}
\label{alg:application}
\DontPrintSemicolon
\KwIn{$D$: target document; $M$: reference model; linguistic-core parameters.}
\KwOut{$\Lambda(D,SRD)$, $\lambda(D,SRD)$, $\lambda^{\mathrm{rel}}(D,SRD)$.}
\BlankLine
$U(D) \leftarrow \mathrm{Segmentation}(D)$ \tcp*{sentences $S_1,\dots,S_n$}
\ForEach{$S_i \in U(D)$}{
  $\mathcal{T}(S_i) \leftarrow \mathrm{LinguisticCore}(S_i)$ \tcp*{steps 1--5 of Alg.~\ref{alg:training}}
  \ForEach{concept $c_j$}{
    $\Lambda_{ij} \leftarrow \mathrm{UnitAffinity}(\mathcal{T}(S_i), M, c_j)$ \tcp*{Eq.~\ref{eq:unit-level-affinity}}
  }
}
$\lambda \leftarrow \mathcal{A}\big(\Lambda\big)$ \tcp*{aggregation per concept (Eq.~\ref{eq:doc-level-affinity})}
$\lambda^{\mathrm{rel}} \leftarrow \lambda \,/\, \lambda^{\mathrm{ref}}$ \tcp*{Eq.~\ref{eq:relative-affinity}}
\Return $\Lambda,\ \lambda,\ \lambda^{\mathrm{rel}}$\;
\end{algorithm}

 
\section{Consistency Study: Sustainable Development Goals}\label{sec:case}

The methodology described in Section~\ref{sec:methodology} is independent of the specific domain, language, and corpus. Both linguistic processing and the computation of the weighting measures can therefore be applied to any $SRD$ and target document, regardless of the language in which they are written. 

This section presents a reference instantiation based on the Sustainable Development Goals (SDGs) of the 2030 Agenda of the United Nations \cite{UN2030Agenda2015}, which provided the initial motivation for this research within the project \emph{Universities for Sustainable Development} mentioned above. In addition to illustrating the complete instantiation of the method, this consistency study constitutes an internal consistency assessment of ITL conducted in this work. Specifically, the SDG statements are evaluated against the reference framework that they themselves define, providing a controlled, self-referential setting in which each statement is associated with a clearly dominant concept by construction. This setting enables the assessment of whether the method separates the induced concept profiles by assigning each statement its highest affinity to the corresponding concept.

The parameters and components specific to this instantiation are summarized in Table~\ref{tab:parameters}.

\subsection{The SDG SRD}
\label{subsec:case-srd}

The SDGs constitute a concept-structured reference framework suitable for instantiating the $SRD$. The framework consists of $17$ goals, each accompanied by a descriptive statement. In this instantiation, each goal corresponds to a concept $c_j$ in the $SRD$, and its official statement serves as descriptor $R_j$, so that the concept set is $\mathcal{C}=\{c_1,\dots,c_{17}\}$ and $m=17$. The complete official statement of each of the $17$ goals, used as its descriptor, is provided in Appendix~\ref{sec:appendix-sdg}. The target document evaluated in this consistency study is the set of SDG statements itself. This configuration enables analysis of the method under controlled conditions in which each textual unit is associated with a known dominant concept.

For each descriptor $R_j$, the linguistic-processing core described in Section~\ref{subsubsec:nlp-core} extracts the reference term set $RTS_j$ and its four term families. As an example, Table~\ref{tab:terms-types} presents the result for Goal G10: \textit{"Reduce inequality within and among countries"}. From three independent terms (\texttt{reduc}\{VB\}, \texttt{inequ}\{NN\}, and \texttt{countri}\{NNS\}), the corresponding bigrams, trigram, and co-occurrences are derived. This example illustrates two properties of the instantiation. First, the part-of-speech tag forms part of the identity of independent terms. Second, co-occurrences are represented as ordered pairs (e.g., \texttt{countri inequ} and \texttt{inequ reduc}), thereby distinguishing them from direct-adjacency bigrams.

\begin{table}[htbp]
    \centering
    \caption{Reference Term Set for G10: "Reduce inequality within and among countries".}
    
    \label{tab:terms-types}
    \begin{tabular}{ll}
        \toprule
        \textbf{Term} & \textbf{Type} \\
        \midrule
        reduc                  & IND - VB  \\
        inequ                  & IND - NN \\
        countri                & IND - NNS\\
        inequ countri          & 2G  \\
        reduc inequ            & 2G  \\
        reduc inequ countri    & 3G  \\
        countri inequ          & COC \\
        countri reduc          & COC \\
        inequ reduc            & COC \\
        reduc countri          & COC \\
        \bottomrule
    \end{tabular}
\end{table}

\subsection{Instantiation of the Linguistic-Processing Core}
\label{subsec}

Table~\ref{tab:parameters} summarizes the configuration of the reference instantiation. Processing is performed in English, which is the language of the official SDG statements. Sentence segmentation and tokenization are performed using standard natural language processing tools. Part-of-speech tagging uses the \emph{Penn Treebank} tagset, retaining categories that convey conceptual content, namely nouns, verbs, adjectives, adverbs, and their variants. Normalization is performed using the \emph{Snowball (Porter2) stemmer}, and the \emph{stopword} list was generated with KNIME and supplied to the algorithm. Target documents provided in unstructured format undergo a preliminary PDF text-extraction stage using \emph{pdfplumber}.

\begin{table}[h]
\centering
\caption{Parameters and components of the reference instantiation on the SDGs.}
\label{tab:parameters}
\begin{tabular}{lll}
\toprule
\textbf{Element} & \textbf{Choice} & \textbf{Version / Reference} \\
\midrule
Language & English & --- \\
Number of concepts ($m$) & 17 (goals) & --- \\
Runtime & Python & 3.9 \\
NLP toolkit & NLTK & 3.9.1~\cite{nltk} \\
Sentence segmentation & Punkt tokenizer & \cite{punkt} \\
Tokenization & Treebank word tokenizer & \cite{marcus1993penn} \\
POS tagging & Stanford POS Tagger & 4.2.0~\cite{toutanova2003} \\
POS model & \texttt{english-caseless-left3words-distsim} & \cite{toutanova2003} \\
Tagset & Penn Treebank & \cite{marcus1993penn} \\
Retained categories & Nouns, verbs, adjectives, adverbs & --- \\
Normalization & Snowball (Porter2) stemmer, English & \cite{porter2snowball} \\
Stopword list & KNIME English stopword list & --- \\
PDF text extraction & pdfplumber & 0.11.4~\cite{pdfplumber} \\
\texttt{min\_short\_len} & 3 & --- \\
\texttt{n\_sw\_coo} ($N_{sw}$) & 5 & --- \\
\bottomrule
\end{tabular}
\end{table}

\subsection{Term Weighting}
\label{subsec:case-weighting}

The weighting measures defined in Section~\ref{subsec:disc-spec-rel-imp} are computed over the constructed $SRD$. Table~\ref{tab:frequency-discriminability} presents an example for a selected set of terms, reporting their frequency in each goal and their discriminability $\delta$. Terms occurring in a limited subset of goals obtain discriminability values closer to 1 (e.g., \texttt{promot inclus}), with maximum discriminability ($\delta=1$) attained when a term occurs in a single goal. Terms shared across several goals, by contrast, obtain lower discriminability values. More transversal terms (e.g., \texttt{sustain}) exhibit lower discriminability, consistent with their reduced capacity to distinguish a specific concept.

\begin{table}[htbp]
    \centering
    \caption{Frequency and discriminability of terms}
    \label{tab:frequency-discriminability}
    
    \resizebox{\textwidth}{!}{%
        \begin{tabular}{ll*{17}{c}c}
            \toprule
            & & \multicolumn{17}{c}{\textbf{Frequency}} & \\
            \cmidrule(lr){3-19}
            \textbf{Term}
            & \textbf{Type}
            & \textbf{G1}
            & \textbf{G2}
            & \textbf{G3}
            & \textbf{G4}
            & \textbf{G5}
            & \textbf{G6}
            & \textbf{G7}
            & \textbf{G8}
            & \textbf{G9}
            & \textbf{G10}
            & \textbf{G11}
            & \textbf{G12}
            & \textbf{G13}
            & \textbf{G14}
            & \textbf{G15}
            & \textbf{G16}
            & \textbf{G17}
            & \textbf{Discriminability} \\
            \midrule
            
            promot inclus
            & COC
            &   &   &   & 1 &   &   &   & 1
            &   &   &   &   &   &   &   & 1 &  
            & 0.612239 \\
            
            sustain
            & IND - JJ
            &   & 1 &   &   &   & 1 & 1 & 2
            & 1 &   & 1 & 1 &   & 1 & 1 & 1 & 1
            & 0.163710 \\
            
            ensur
            & IND - VB
            &   &   & 1 & 1 &   & 1 & 1 &
            &   &   &   & 1 &   &   &   &   &
            & 0.431939 \\
            
            \bottomrule
        \end{tabular}%
    }
    
\end{table}

Table~\ref{tab:number-terms} reports the frequency of terms of each type for each goal, which determines the size of the corresponding pattern space and, consequently, the specificity. Table~\ref{tab:specificity-values} presents the resulting specificity values $\sigma$, computed by term type and separately for each concept. The resulting values exhibit the consistent hierarchy across term types: independent terms, which are more numerous, receive lower specificity, whereas more restrictive compound patterns receive higher values. This gradation reflects the assumption that a contextually richer pattern has a lower probability of occurring by chance and is therefore semantically more specific.

\begin{table}[htbp]
    \centering
    \caption{Frequencies of terms by type and goal}
    \label{tab:number-terms}
    
    \resizebox{\textwidth}{!}{%
        \begin{tabular}{l*{17}{c}}
            \toprule
            \textbf{Type}
            & \textbf{G1}
            & \textbf{G2}
            & \textbf{G3}
            & \textbf{G4}
            & \textbf{G5}
            & \textbf{G6}
            & \textbf{G7}
            & \textbf{G8}
            & \textbf{G9}
            & \textbf{G10}
            & \textbf{G11}
            & \textbf{G12}
            & \textbf{G13}
            & \textbf{G14}
            & \textbf{G15}
            & \textbf{G16}
            & \textbf{G17} \\
            \midrule
            
            IND
            & 2 & 9 & 5 & 9 & 6 & 6 & 7 & 8 & 9
            & 3 & 7 & 5 & 6 & 8 & 17 & 14 & 8 \\
            
            2G
            & 1 & 8 & 4 & 8 & 5 & 5 & 6 & 8 & 8
            & 2 & 6 & 4 & 5 & 7 & 17 & 14 & 7 \\
            
            3G
            & 0 & 7 & 3 & 7 & 4 & 4 & 5 & 7 & 7
            & 1 & 5 & 3 & 4 & 6 & 16 & 13 & 6 \\
            
            COC
            & 1 & 52 & 16 & 52 & 25 & 25 & 34 & 40 & 52
            & 4 & 34 & 16 & 25 & 33 & 107 & 104 & 43 \\
            
            \bottomrule
        \end{tabular}%
    }
\end{table}

\begin{table}[htbp]
    \centering
    \caption{Term-type specificity weights by Sustainable Development Goal.}
    \label{tab:specificity-values}
    
    \resizebox{\textwidth}{!}{%
        \begin{tabular}{l*{17}{c}}
            \toprule
            \textbf{Type}
            & \textbf{G1}
            & \textbf{G2}
            & \textbf{G3}
            & \textbf{G4}
            & \textbf{G5}
            & \textbf{G6}
            & \textbf{G7}
            & \textbf{G8}
            & \textbf{G9}
            & \textbf{G10}
            & \textbf{G11}
            & \textbf{G12}
            & \textbf{G13}
            & \textbf{G14}
            & \textbf{G15}
            & \textbf{G16}
            & \textbf{G17} \\
            \midrule
            
            IND
            & 0.500000
            & 0.049724
            & 0.121951
            & 0.049724
            & 0.090909
            & 0.090909
            & 0.072165
            & 0.060606
            & 0.049724
            & 0.272727
            & 0.072165
            & 0.121951
            & 0.090909
            & 0.064000
            & 0.018085
            & 0.024433
            & 0.059259 \\
            
            2G
            & 0.250000
            & 0.198895
            & 0.243902
            & 0.198895
            & 0.227273
            & 0.227273
            & 0.216495
            & 0.212121
            & 0.198895
            & 0.272727
            & 0.216495
            & 0.243902
            & 0.227273
            & 0.224000
            & 0.144681
            & 0.158813
            & 0.207407 \\
            
            3G
            & 0.000000
            & 0.464088
            & 0.243902
            & 0.464088
            & 0.303030
            & 0.303030
            & 0.360825
            & 0.424242
            & 0.464088
            & 0.090909
            & 0.360825
            & 0.243902
            & 0.303030
            & 0.448000
            & 0.723404
            & 0.635253
            & 0.414815 \\
            
            COC
            & 0.250000
            & 0.287293
            & 0.390244
            & 0.287293
            & 0.378788
            & 0.378788
            & 0.350515
            & 0.303030
            & 0.287293
            & 0.363636
            & 0.350515
            & 0.390244
            & 0.378788
            & 0.264000
            & 0.113830
            & 0.181501
            & 0.318519 \\
            
            \bottomrule
        \end{tabular}%
    }
\end{table}

The combination of relevance and discriminability determines the importance $\varphi$ of each term with respect to each goal and therefore defines the reference model $M$. Table~\ref{tab:frequency-discriminability-importance} illustrates the portion of the reference model associated with the terms presented in Table~\ref{tab:frequency-discriminability}. Importance consequently integrates the two dimensions of the model: a term attains high importance for a goal when it belongs to the corresponding terminological profile with high specificity and simultaneously discriminates that goal from the remaining goals.

Finally, the concept-specific reference affinity is computed. This value serves as the interpretive anchor during the document-alignment phase, thereby completing the modeling of the $SRD$ and the reference-model induction phase.

\begin{table}[h]
	\centering
	\caption{Importance of terms distributed among goals.}
	\label{tab:frequency-discriminability-importance}
	
	\setlength{\tabcolsep}{2.5pt}
	\renewcommand{\arraystretch}{1.2}
	
	\resizebox{\linewidth}{!}{%
		\begin{tabular}{ll*{17}{c}}
			\toprule
			\textbf{Term} & \textbf{Type}
			& \textbf{G1} & \textbf{G2} & \textbf{G3} & \textbf{G4} & \textbf{G5} & \textbf{G6} & \textbf{G7} & \textbf{G8} & \textbf{G9} & \textbf{G10} & \textbf{G11} & \textbf{G12} & \textbf{G13} & \textbf{G14} & \textbf{G15} & \textbf{G16} & \textbf{G17} \\
			\midrule
			promot inclus & COC
			& 0 & 0 & 0 & 0.175891979 & 0 & 0 & 0 & 0.185526784 & 0 & 0 & 0 & 0 & 0 & 0 & 0 & 0.111121991 & 0 \\
			sustain & IND - JJ
			& 0 & 0.008140316 & 0 & 0 & 0 & 0.014882712 & 0.011814132 & 0.009921808 & 0.008140316 & 0 & 0.011814132 & 0.019964598 & 0 & 0.010477440 & 0.002960695 & 0.003999926 & 0.009701291 \\
			ensur & IND - VB
			& 0 & 0 & 0.052675393 & 0.021477735 & 0 & 0.039267143 & 0.031170878 & 0 & 0 & 0 & 0 & 0.052675393 & 0 & 0 & 0 & 0 & 0 \\
			\bottomrule
		\end{tabular}%
	}
\end{table}

\subsection{Document-Alignment Phase and Affinity Computation}
\label{subsec:case-affinity}

To illustrate the alignment phase and affinity computation, the SDG statements are evaluated against the reference framework that they themselves define. This evaluation produces the affinity matrix shown in Table~\ref{tab:affinity-matrix}, in which each cell represents the affinity of the statement of goal $i$ with respect to concept $j$. The resulting matrix exhibits a dominant diagonal: each statement attains its maximum affinity with the corresponding concept, whereas off-diagonal values are substantially lower. The diagonal corresponds to the \emph{reference affinity} $\lambda^{\mathrm{ref}}_j(SRD)$, that is, the affinity of each statement with itself.

The diagonal affinity values range from $0.18$ to $0.38$, depending on the goal, whereas cross-affinity between distinct goals is close to zero. The mean off-diagonal affinity is two orders of magnitude lower than the mean diagonal affinity, and the largest observed cross-affinity value ($0.020$) remains below the lowest reference affinity ($0.18$). This separation indicates that the method distinguishes among the $17$ goals by assigning each statement its maximum affinity to the corresponding dominant concept, with limited cross-affinity among the remaining concepts.

\begin{table}[h]
\centering
\caption{Affinity matrix}
\label{tab:affinity-matrix}

\scriptsize
\setlength{\tabcolsep}{2pt}
\renewcommand{\arraystretch}{1.15}
\arrayrulecolor{gray!60}

\resizebox{\linewidth}{!}{%
\begin{tabular}{|L|*{17}{C|}}
    \hline
    \multicolumn{1}{|c|}{\cellcolor{white}} &
    \bfseries G1 &
    \bfseries G2 &
    \bfseries G3 &
    \bfseries G4 &
    \bfseries G5 &
    \bfseries G6 &
    \bfseries G7 &
    \bfseries G8 &
    \bfseries G9 &
    \bfseries G10 &
    \bfseries G11 &
    \bfseries G12 &
    \bfseries G13 &
    \bfseries G14 &
    \bfseries G15 &
    \bfseries G16 &
    \bfseries G17 \\
    \hline

    \rowhdr{G1}  & \databar{0.375}    & \databar{0}        & \databar{0}        & \databar{0}        & \databar{0}        & \databar{0}        & \databar{0}        & \databar{0}        & \databar{0}        & \databar{0}        & \databar{0}        & \databar{0}        & \databar{0}        & \databar{0}        & \databar{0}        & \databar{0}        & \databar{0} \\
    \hline
    \rowhdr{G2}  & \databar{0}        & \databar{0.261899} & \databar{0.000503} & \databar{0.000205} & \databar{0.000904} & \databar{0.000196} & \databar{0.000155} & \databar{0.003895} & \databar{0.002024} & \databar{0}        & \databar{0.000155} & \databar{0.000263} & \databar{0}        & \databar{0.000138} & \databar{0.001957} & \databar{0.001235} & \databar{0.000128} \\
    \hline
    \rowhdr{G3}  & \databar{0}        & \databar{0.000556} & \databar{0.293464} & \databar{0.016824} & \databar{0}        & \databar{0.001402} & \databar{0.001113} & \databar{0.000678} & \databar{0.000556} & \databar{0}        & \databar{0}        & \databar{0.001881} & \databar{0}        & \databar{0}        & \databar{0.000202} & \databar{0.000273} & \databar{0} \\
    \hline
    \rowhdr{G4}  & \databar{0}        & \databar{0.000205} & \databar{0.008953} & \databar{0.259792} & \databar{0}        & \databar{0.000517} & \databar{0.000410} & \databar{0.005085} & \databar{0.002429} & \databar{0}        & \databar{0.000426} & \databar{0.000693} & \databar{0}        & \databar{0}        & \databar{0.000075} & \databar{0.002927} & \databar{0} \\
    \hline
    \rowhdr{G5}  & \databar{0}        & \databar{0.000939} & \databar{0}        & \databar{0}        & \databar{0.308535} & \databar{0}        & \databar{0}        & \databar{0}        & \databar{0}        & \databar{0}        & \databar{0}        & \databar{0}        & \databar{0}        & \databar{0}        & \databar{0}        & \databar{0}        & \databar{0} \\
    \hline
    \rowhdr{G6}  & \databar{0}        & \databar{0.000204} & \databar{0.001317} & \databar{0.000537} & \databar{0}        & \databar{0.296393} & \databar{0.013059} & \databar{0.000248} & \databar{0.000204} & \databar{0}        & \databar{0.000295} & \databar{0.007789} & \databar{0}        & \databar{0.000262} & \databar{0.005013} & \databar{0.000100} & \databar{0.000243} \\
    \hline
    \rowhdr{G7}  & \databar{0}        & \databar{0.000157} & \databar{0.001013} & \databar{0.000413} & \databar{0}        & \databar{0.011003} & \databar{0.288723} & \databar{0.000191} & \databar{0.000157} & \databar{0}        & \databar{0.000227} & \databar{0.005992} & \databar{0}        & \databar{0.000201} & \databar{0.000057} & \databar{0.005705} & \databar{0.000187} \\
    \hline
    \rowhdr{G8}  & \databar{0}        & \databar{0.006570} & \databar{0.000606} & \databar{0.005723} & \databar{0}        & \databar{0.000472} & \databar{0.000375} & \databar{0.285260} & \databar{0.009705} & \databar{0}        & \databar{0.000890} & \databar{0.019845} & \databar{0}        & \databar{0.000333} & \databar{0.003227} & \databar{0.006268} & \databar{0.000308} \\
    \hline
    \rowhdr{G9}  & \databar{0}        & \databar{0.002024} & \databar{0.000503} & \databar{0.002429} & \databar{0}        & \databar{0.000196} & \databar{0.000155} & \databar{0.006689} & \databar{0.251039} & \databar{0}        & \databar{0.014574} & \databar{0.000263} & \databar{0}        & \databar{0.000138} & \databar{0.000792} & \databar{0.013324} & \databar{0.000128} \\
    \hline
    \rowhdr{G10} & \databar{0}        & \databar{0}        & \databar{0}        & \databar{0}        & \databar{0}        & \databar{0}        & \databar{0}        & \databar{0}        & \databar{0}        & \databar{0.290909} & \databar{0}        & \databar{0}        & \databar{0}        & \databar{0}        & \databar{0}        & \databar{0}        & \databar{0} \\
    \hline
    \rowhdr{G11} & \databar{0}        & \databar{0.000157} & \databar{0}        & \databar{0.000429} & \databar{0}        & \databar{0.000286} & \databar{0.000227} & \databar{0.000714} & \databar{0.017210} & \databar{0}        & \databar{0.287097} & \databar{0.000384} & \databar{0}        & \databar{0.000201} & \databar{0.000057} & \databar{0.005061} & \databar{0.000187} \\
    \hline
    \rowhdr{G12} & \databar{0}        & \databar{0.000291} & \databar{0.001881} & \databar{0.000767} & \databar{0}        & \databar{0.010216} & \databar{0.009199} & \databar{0.017137} & \databar{0.000291} & \databar{0}        & \databar{0.000422} & \databar{0.287966} & \databar{0}        & \databar{0.000374} & \databar{0.000106} & \databar{0.000143} & \databar{0.000346} \\
    \hline
    \rowhdr{G13} & \databar{0}        & \databar{0}        & \databar{0}        & \databar{0}        & \databar{0}        & \databar{0}        & \databar{0}        & \databar{0}        & \databar{0}        & \databar{0}        & \databar{0}        & \databar{0}        & \databar{0.309091} & \databar{0}        & \databar{0}        & \databar{0}        & \databar{0} \\
    \hline
    \rowhdr{G14} & \databar{0}        & \databar{0.000151} & \databar{0}        & \databar{0}        & \databar{0}        & \databar{0.000276} & \databar{0.000219} & \databar{0.000184} & \databar{0.000151} & \databar{0}        & \databar{0.000219} & \databar{0.000370} & \databar{0}        & \databar{0.273718} & \databar{0.000308} & \databar{0.004209} & \databar{0.006814} \\
    \hline
    \rowhdr{G15} & \databar{0}        & \databar{0.002584} & \databar{0.000243} & \databar{0.000099} & \databar{0}        & \databar{0.004929} & \databar{0.000075} & \databar{0.002760} & \databar{0.001809} & \databar{0}        & \databar{0.000075} & \databar{0.000127} & \databar{0}        & \databar{0.000375} & \databar{0.177286} & \databar{0.001122} & \databar{0.000062} \\
    \hline
    \rowhdr{G16} & \databar{0}        & \databar{0.001061} & \databar{0.000263} & \databar{0.002640} & \databar{0}        & \databar{0.000103} & \databar{0.004109} & \databar{0.003868} & \databar{0.011336} & \databar{0}        & \databar{0.003834} & \databar{0.000138} & \databar{0}        & \databar{0.002403} & \databar{0.000415} & \databar{0.201280} & \databar{0.002538} \\
    \hline
    \rowhdr{G17} & \databar{0}        & \databar{0.000127} & \databar{0}        & \databar{0}        & \databar{0}        & \databar{0.000233} & \databar{0.000185} & \databar{0.000155} & \databar{0.000127} & \databar{0}        & \databar{0.000185} & \databar{0.000312} & \databar{0}        & \databar{0.005444} & \databar{0.000046} & \databar{0.003552} & \databar{0.278667} \\
    \hline
\end{tabular}%
}

\arrayrulecolor{black}
\end{table}

Differences in reference affinity across goals reflect differences in the terminological richness of their statements. Goals with longer descriptors and a greater presence of specific compound patterns tend to attain higher reference affinities, whereas shorter statements obtain lower values. These reference values provide the interpretive anchor used to normalize the affinity of external target documents during the alignment phase. This means that relative affinity expresses document alignment on a scale defined with respect to the corresponding concept descriptor.

Overall, this experiment provides an internal consistency assessment of ITL on the reference descriptors from which the model is induced. Because the evaluation units are identical to the concept descriptors used to construct the $SRD$, this experiment does not assess generalization to unseen wording, paraphrases, or external documents. ITL does not perform exclusive concept assignment. For each textual unit, it instead quantifies affinity with each concept in the $SRD$. A single unit may therefore align simultaneously with several concepts when supported by its content, with those concepts receiving higher affinity values than the remaining concepts.

In this consistency study, the characteristics of the corpus, that is, official statements evaluated against the framework that they themselves define, result in each statement having a clearly dominant concept. This property provides a known reference against which the results can be assessed. The method assigns each statement its maximum affinity to the corresponding concept and separates this affinity from the marginal affinities associated with the remaining concepts. Within this experimental setting, the result indicates that the weighting measures (discriminability, specificity, and importance) and their integration into the affinity score effectively capture the relationship between a document and a concept-structured reference framework.

The evaluation of ITL on heterogeneous external documents, in which a textual unit may align simultaneously with multiple concepts, as well as its systematic comparison with other semantic-alignment approaches, remains outside the scope of this work and is proposed as future research.

In this consistency study, affinity is evaluated at the textual-unit level; therefore, neither document-level aggregated affinities nor the absolute (Equation~\ref{eq:doc-affinity-vector}) and relative (Equation~\ref{eq:relative-affinity}) affinity vectors are computed. The reason is that the evaluated set does not constitute a cohesive document, but rather a collection of $17$ independent statements, each associated with a distinct concept. Aggregating their affinities into a single vector would therefore combine heterogeneous content units and produce a concept-level score without a meaningful interpretation. By contrast, document-level aggregation is applicable when the target document is a single extended text whose units share a common thematic structure, as in the alignment phase involving external documents.

\section{Conclusions}
\label{sec:conclusions}
This work has presented \emph{Intelligent Target Locator} (ITL), a methodology for quantifying, in an interpretable and traceable manner, the alignment between a target document and a \emph{Structured Reference Document} ($SRD$). Its central contribution is the formalization of an original affinity metric that provides a graded measure for each \emph{textual unit, concept} pair. To this end, ITL directly induces from the $SRD$ a reference model composed of concept-specific terminological profiles that integrate independent terms, bigrams, trigrams, and co-occurrences. The contribution of each term is determined by combining its frequency in the evaluated unit with its importance for the concept. This importance is defined according to the term's membership in the terminological profile, its term-type-dependent specificity, and its discriminability based on the distribution of the term across the concepts in the $SRD$. Aggregating these contributions produces a textual-unit--concept affinity matrix that preserves traceability to the terms supporting each value. In addition, the reference affinity of a concept with itself provides a concept-specific interpretive anchor, while aggregation operators extend the analysis to higher levels of granularity.

An internal consistency assessment using the 17 SDG descriptors showed that each descriptor received its highest affinity with its corresponding concept. Because these evaluation units also define the SRD, the experiment demonstrates separation of the induced concept profiles rather than generalization to unseen text. Moreover, the mean off-diagonal affinity was two orders of magnitude lower than the mean reference affinity, and the largest cross-affinity remained below the lowest reference affinity. These results support the ability of the affinity metric to distinguish among the conceptual profiles defined by the $SRD$ and the coherence of the proposed weighting scheme.

The architecture is portable in principle to other structured reference frameworks because its concept profiles are induced directly from the reference document. Empirical generalization across domains and languages remains to be established. The methodology could similarly be instantiated for other normative, strategic, regulatory, or programmatic frameworks and adapted to different languages through the corresponding linguistic-processing resources.

This independence from domain and language, together with the traceability of the affinity values, extends its applicability to scenarios requiring the quantification and explanation of the alignment of heterogeneous documentation with an explicit reference framework.

Future work will extend empirical validation to external documents, additional domains, and languages, examine alternative document-level aggregation strategies, and comparatively evaluate ITL against other semantic-representation and alignment approaches.


\section*{Acknowledgements}
This work was funded by Grant PID2023-152660NB-I00 funded by the Spanish Ministry of Science, Innovation and Universities.

\bibliographystyle{unsrt}  


\bibliography{bibliography}

@misc{UN2030Agenda2015,
  author       = {{United Nations General Assembly}},
  title        = {Transforming our world: the 2030 Agenda for Sustainable Development},
  howpublished = {Resolution A/RES/70/1},
  year         = {2015},
  url          = {https://undocs.org/A/RES/70/1},
  note         = {Accessed: 2026-01-15}
}

@article{RobertsonZaragoza2009,
  author  = {Robertson, Stephen E. and Zaragoza, Hugo},
  title   = {The Probabilistic Relevance Framework: {BM25} and Beyond},
  journal = {Foundations and Trends in Information Retrieval},
  year    = {2009},
  volume  = {3},
  number  = {4},
  pages   = {333--389},
  doi     = {10.1561/1500000019},
  url     = {https://dl.acm.org/doi/abs/10.1561/1500000019}
}

@inproceedings{Devlin2019BERT,
  author    = {Devlin, Jacob and Chang, Ming-Wei and Lee, Kenton and Toutanova, Kristina},
  title     = {{BERT}: Pre-training of Deep Bidirectional Transformers for Language Understanding},
  booktitle = {Proceedings of the 2019 Conference of the North American Chapter of the Association for Computational Linguistics: Human Language Technologies (Volume 1: Long and Short Papers)},
  year      = {2019},
  pages     = {4171--4186},
  address   = {Minneapolis, Minnesota},
  publisher = {Association for Computational Linguistics},
  doi       = {10.18653/v1/N19-1423},
  url       = {https://aclanthology.org/N19-1423/}
}

@inproceedings{Reimers2019SBERT,
  author    = {Reimers, Nils and Gurevych, Iryna},
  title     = {Sentence-{BERT}: Sentence Embeddings using Siamese {BERT}-Networks},
  booktitle = {Proceedings of the 2019 Conference on Empirical Methods in Natural Language Processing and the 9th International Joint Conference on Natural Language Processing ({EMNLP}-{IJCNLP})},
  year      = {2019},
  pages     = {3982--3992},
  address   = {Hong Kong, China},
  publisher = {Association for Computational Linguistics},
  doi       = {10.18653/v1/D19-1410},
  url       = {https://aclanthology.org/D19-1410/}
}

@inproceedings{Karpukhin2020DPR,
  author    = {Karpukhin, Vladimir and Oguz, Barlas and Min, Sewon and Lewis, Patrick and Wu, Ledell and Edunov, Sergey and Chen, Danqi and Yih, Wen-tau},
  title     = {Dense Passage Retrieval for Open-Domain Question Answering},
  booktitle = {Proceedings of the 2020 Conference on Empirical Methods in Natural Language Processing ({EMNLP})},
  year      = {2020},
  pages     = {6769--6781},
  address   = {Online},
  publisher = {Association for Computational Linguistics},
  doi       = {10.18653/v1/2020.emnlp-main.550},
  url       = {https://aclanthology.org/2020.emnlp-main.550/}
}

@inproceedings{KhattabZaharia2020ColBERT,
  author    = {Khattab, Omar and Zaharia, Matei},
  title     = {Col{BERT}: Efficient and Effective Passage Search via Contextualized Late Interaction over {BERT}},
  booktitle = {Proceedings of the 43rd International {ACM} {SIGIR} Conference on Research and Development in Information Retrieval ({SIGIR} '20)},
  year      = {2020},
  pages     = {39--48},
  publisher = {ACM},
  doi       = {10.1145/3397271.3401075},
  url       = {https://dblp.org/rec/conf/sigir/KhattabZ20}
}

@article{Fotopoulou2022SustainGraph,
  author  = {Fotopoulou, Eleni and Mandilara, Ioanna and Zafeiropoulos, Anastasios and Laspidou, Chrysi and Adamos, Giannis and Koundouri, Phoebe and Papavassiliou, Symeon},
  title   = {SustainGraph: A knowledge graph for tracking the progress and the interlinking among the sustainable development goals' targets},
  journal = {Frontiers in Environmental Science},
  year    = {2022},
  volume  = {10},
  pages   = {1003599},
  doi     = {10.3389/fenvs.2022.1003599},
  url     = {https://doi.org/10.3389/fenvs.2022.1003599}
}

@inproceedings{MihalceaTarau2004TextRank,
  author    = {Mihalcea, Rada and Tarau, Paul},
  title     = {TextRank: Bringing Order into Text},
  booktitle = {Proceedings of the 2004 Conference on Empirical Methods in Natural Language Processing},
  year      = {2004},
  pages     = {404--411},
  address   = {Barcelona, Spain},
  publisher = {Association for Computational Linguistics},
  url       = {https://aclanthology.org/W04-3252/}
}

@article{Campos2020YAKE,
  author  = {Campos, Ricardo and Mangaravite, V{\'i}tor and Pasquali, Arian and Jorge, Al{\'i}pio and Nunes, C{\'e}lia and Jatowt, Adam},
  title   = {{YAKE}! Keyword extraction from single documents using multiple local features},
  journal = {Information Sciences},
  year    = {2020},
  volume  = {509},
  pages   = {257--289},
  doi     = {10.1016/j.ins.2019.09.013},
  url     = {https://doi.org/10.1016/j.ins.2019.09.013}
}

@article{JustesonKatz1995TechnicalTerminology,
  author  = {Justeson, John S. and Katz, Slava M.},
  title   = {Technical terminology: some linguistic properties and an algorithm for identification in text},
  journal = {Natural Language Engineering},
  year    = {1995},
  volume  = {1},
  number  = {1},
  pages   = {9--27},
  doi     = {10.1017/S1351324900000048},
  url     = {https://doi.org/10.1017/S1351324900000048}
}

@article{Frantzi2000CValue,
  author  = {Frantzi, Katerina T. and Ananiadou, Sophia and Mima, Hideki},
  title   = {Automatic recognition of multi-word terms: the C-value/NC-value method},
  journal = {International Journal on Digital Libraries},
  year    = {2000},
  volume  = {3},
  number  = {2},
  pages   = {115--130},
  doi     = {10.1007/s007999900023},
  url     = {https://doi.org/10.1007/s007999900023}
}

@article{Blei2003LDA,
  author  = {Blei, David M. and Ng, Andrew Y. and Jordan, Michael I.},
  title   = {Latent Dirichlet Allocation},
  journal = {Journal of Machine Learning Research},
  year    = {2003},
  volume  = {3},
  pages   = {993--1022},
  url     = {https://dl.acm.org/doi/10.5555/944919.944937}
}

@book{nltk,
  title     = {Natural Language Processing with Python},
  author    = {Bird, Steven and Klein, Ewan and Loper, Edward},
  year      = {2009},
  publisher = {O'Reilly Media}
}

@article{punkt,
  title   = {Unsupervised Multilingual Sentence Boundary Detection},
  author  = {Kiss, Tibor and Strunk, Jan},
  journal = {Computational Linguistics},
  volume  = {32},
  number  = {4},
  pages   = {485--525},
  year    = {2006}
}

@inproceedings{toutanova2003,
  title     = {Feature-Rich Part-of-Speech Tagging with a Cyclic Dependency Network},
  author    = {Toutanova, Kristina and Klein, Dan and Manning, Christopher D. and Singer, Yoram},
  booktitle = {Proceedings of the 2003 Conference of the North American Chapter of the Association for Computational Linguistics (NAACL-HLT)},
  pages     = {173--180},
  year      = {2003}
}

@article{marcus1993penn,
  title   = {Building a Large Annotated Corpus of English: The Penn Treebank},
  author  = {Marcus, Mitchell P. and Santorini, Beatrice and Marcinkiewicz, Mary Ann},
  journal = {Computational Linguistics},
  volume  = {19},
  number  = {2},
  pages   = {313--330},
  year    = {1993}
}

@misc{porter2snowball,
  title        = {Snowball: A Language for Stemming Algorithms},
  author       = {Porter, Martin F.},
  year         = {2001},
  howpublished = {\url{https://snowballstem.org/}}
}

@misc{pdfplumber,
  title        = {pdfplumber: Plumb a {PDF} for detailed information about each char, rectangle, line and table},
  author       = {Singer-Vine, Jeremy},
  howpublished = {\url{https://github.com/jsvine/pdfplumber}}
}

@inproceedings{Guisiano2022SDGMeter,
  author    = {Guisiano, Jade Eva and Chiky, Raja and De Mello, Jonathas},
  title     = {{SDG-Meter}: A Deep Learning Based Tool for Automatic Text Classification of the Sustainable Development Goals},
  booktitle = {Intelligent Information and Database Systems ({ACIIDS} 2022)},
  series    = {Lecture Notes in Computer Science},
  volume    = {13757},
  pages     = {259--271},
  publisher = {Springer},
  year      = {2022},
  doi       = {10.1007/978-3-031-21743-2_21},
  url       = {https://doi.org/10.1007/978-3-031-21743-2_21}
}

@article{Pukelis2022OSDG,
  author  = {Pukelis, Lukas and Bautista-Puig, Nuria and Statulevi{\v{c}}i{\=u}t{\.e}, Gust{\.e} and Stan{\v{c}}iauskas, Vilius and Dikmener, Gokhan and Akylbekova, Dina},
  title   = {{OSDG} 2.0: A Multilingual Tool for Classifying Text Data by {UN} Sustainable Development Goals ({SDGs})},
  journal = {arXiv preprint arXiv:2211.11252},
  year    = {2022},
  doi     = {10.48550/arXiv.2211.11252},
  url     = {https://arxiv.org/abs/2211.11252}
}

@inproceedings{Skrynnyk2024SDGiCorpus,
  author    = {Skrynnyk, Mykola and Disassa, Gedion and Krachkov, Andrey and DeVera, Janine},
  title     = {{SDGi} Corpus: A Comprehensive Multilingual Dataset for Text Classification by Sustainable Development Goals},
  booktitle = {Proceedings of the 2nd Symposium on {NLP} for Social Good ({NSG} 2024)},
  series    = {CEUR Workshop Proceedings},
  volume    = {3764},
  pages     = {32--42},
  year      = {2024},
  url       = {https://ceur-ws.org/Vol-3764/paper3.pdf}
}

@inproceedings{Joshi2021SDGKOS,
  author    = {Joshi, Amit and Gonz{\'a}lez Morales, Luis G. and Klarman, Szymon and Stellato, Armando and Helton, Aaron and Lovell, Sean and Haczek, Artur},
  title     = {A Knowledge Organization System for the United Nations Sustainable Development Goals},
  booktitle = {The Semantic Web ({ESWC} 2021)},
  series    = {Lecture Notes in Computer Science},
  volume    = {12731},
  pages     = {548--564},
  publisher = {Springer},
  year      = {2021},
  doi       = {10.1007/978-3-030-77385-4_33},
  url       = {https://doi.org/10.1007/978-3-030-77385-4_33}
}


\newpage
\appendix
\section{The 17 Sustainable Development Goals}
\label{sec:appendix-sdg}

The 2030 Agenda for Sustainable Development of the United Nations~\cite{UN2030Agenda2015}
is organized into 17 goals. The official statement of each goal, used verbatim as the
concept descriptor $R_j$ in the reference instantiation of Section~\ref{sec:case}, is
listed below.

\begin{enumerate}
  \item End poverty in all its forms everywhere.
  \item End hunger, achieve food security and improved nutrition and promote sustainable agriculture.
  \item Ensure healthy lives and promote well-being for all at all ages.
  \item Ensure inclusive and equitable quality education and promote lifelong learning opportunities for all.
  \item Achieve gender equality and empower all women and girls.
  \item Ensure availability and sustainable management of water and sanitation for all.
  \item Ensure access to affordable, reliable, sustainable and modern energy for all.
  \item Promote sustained, inclusive and sustainable economic growth, full and productive employment and decent work for all.
  \item Build resilient infrastructure, promote inclusive and sustainable industrialization and foster innovation.
  \item Reduce inequality within and among countries.
  \item Make cities and human settlements inclusive, safe, resilient and sustainable.
  \item Ensure sustainable consumption and production patterns.
  \item Take urgent action to combat climate change and its impacts.
  \item Conserve and sustainably use the oceans, seas and marine resources for sustainable development.
  \item Protect, restore and promote sustainable use of terrestrial ecosystems, sustainably manage forests, combat desertification, and halt and reverse land degradation and halt biodiversity loss.
  \item Promote peaceful and inclusive societies for sustainable development, provide access to justice for all and build effective, accountable and inclusive institutions at all levels.
  \item Strengthen the means of implementation and revitalize the global partnership for sustainable development.
\end{enumerate}

\section{Notation Glossary}\label{sec:glossary}

The symbols, operators, and acronyms used throughout the article are summarized below and grouped according to their role within the formalization. The equation or section in which each element is introduced is indicated in parentheses where applicable.

\paragraph{Documents and Textual Units.}
\begin{itemize}
\item $SRD$: \emph{Structured Reference Document}; encodes the target semantic domain against which alignment is assessed (Section~\ref{subsec:problem-statement}).
\item $\mathcal{C}=\{c_1,\dots,c_m\}$: finite set of concepts encoded in the $SRD$, where $c_j$ denotes the $j$-th concept, with $j\in\{1,\dots,m\}$.
\item $m$: number of concepts in the $SRD$.
\item $R_j$: reference textual unit (definition, paragraph, or text fragment) describing concept $c_j$.
\item $U(SRD)=\{R_1,\dots,R_m\}$: set of reference textual units in the $SRD$.
\item $D$: target document whose conceptual alignment with the $SRD$ is to be quantified.
\item $U(D)=\{S_1,\dots,S_n\}$: ordered sequence of textual units in $D$ (e.g., sentences, segments, or paragraphs) obtained through segmentation.
\item $S_i$: $i$-th textual unit of document $D$, with $i\in\{1,\dots,n\}$.
\item $n$: number of textual units in the target document $D$.
\item $X$: generic textual unit, $X \in U(D)\cup U(SRD)$, used to define term extraction uniformly.
\item $D^{\mathrm{ref}}_j$: reference document associated with concept $c_j$, consisting of a single textual unit, $U(D^{\mathrm{ref}}_j)=\{R_j\}$ (Equation~\ref{eq:ref-document}).
\end{itemize}

\paragraph{Terms, Vocabularies, and Terminological Profiles.}
\begin{itemize}
\item $t$: term; lexical unit resulting from linguistic preprocessing, which may be simple (unigram) or compound ($n$-gram or co-occurrence).
\item $\mathcal{T}(X)$: multiset (set with multiplicity) of terms extracted from textual unit $X$ (Equation~\ref{eq:term-multiset-union}).
\item $f(t\mid X)$: absolute frequency of term $t$ in $X$, that is, its multiplicity in $\mathcal{T}(X)$ (Equation~\ref{eq:term-frequency}).
\item $\mathcal{V}(X)$: term vocabulary of $X$; set of distinct terms in $\mathcal{T}(X)$, disregarding multiplicity (Equation~\ref{eq:unique-vocab}).
\item $RTS_j=\mathcal{T}(R_j)$: \emph{Reference Term Set}, or terminological profile, associated with concept $c_j$ (Equation~\ref{eq:rts-definition}).
\item $\mathcal{TP}=\{RTS_1,\dots,RTS_m\}$: \emph{Terminological-Profile Collection}; collection of the terminological profiles of all concepts in the $SRD$ (Equation~\ref{eq:tp-collection}).
\item $\mathcal{V}_{SRD}$: term universe induced by the $SRD$, defined as the union of the concept-specific vocabularies (Equation~\ref{eq:srd-vocab}).
\item $\mathcal{V}^{\mathrm{IND}}(R_j)$, $\mathcal{V}^{\mathrm{COC}}(R_j)$: type-specific vocabularies of concept $c_j$; the sets of distinct independent terms and of distinct co-occurrences observed in $R_j$, respectively (Equation~\ref{eq:specificity-potentials}).
\end{itemize}

\paragraph{Term Typology.}
\begin{itemize}
\item $w(X)=\{w_1,\dots,w_{L_X}\}$: ordered sequence of normalized \textit{tokens} in textual unit $X$, obtained after filtering and linguistic normalization (Equation~\ref{eq:base-token-seq}).
\item $L_X$: length of $w(X)$, i.e., the number of \textit{tokens} in $X$ after preprocessing.
\item $\uplus$: multiset union operator, which preserves element multiplicities.
\item $\mathcal{T}^{\mathrm{IND}}(X)$: multiset of independent terms, or unigrams, in $X$ (Equation~\ref{eq:ind-terms}).
\item $\mathcal{T}^{\mathrm{2G}}(X)$: multiset of bigrams formed by two consecutive \textit{tokens} in $w(X)$ (Equation~\ref{eq:bigram-terms}).
\item $\mathcal{T}^{\mathrm{3G}}(X)$: multiset of trigrams formed by three consecutive \textit{tokens} in $w(X)$ (Equation~\ref{eq:trigram-terms}).
\item $\mathcal{T}^{\mathrm{COC}}(X)$: multiset of co-occurrences formed by two non-consecutive \textit{tokens} separated by a bounded number of intermediate \textit{tokens} (Equation~\ref{eq:cooc-terms}).
\item $N_{sw}$: maximum number of intermediate \textit{tokens} allowed between the two \textit{tokens} forming a co-occurrence (Equation~\ref{eq:cooc-terms}).
\item $\kappa(t)\in\{\mathrm{IND},\mathrm{2G},\mathrm{3G},\mathrm{COC}\}$: type of term $t$ according to the preceding typology (Section~\ref{subsec:term-typology}).
\item $K=\{\mathrm{IND},\mathrm{2G},\mathrm{3G},\mathrm{COC}\}$: set of term types over which specificity is normalized (Equation~\ref{eq:specificity}).
\item $\ell_k,\,g_k$: stem and part-of-speech tag of \textit{token} $w_k$; the pair $(\ell_k,g_k)$ defines the identity of an independent term (Section~\ref{subsubsec:nlp-core}).
\end{itemize}

\paragraph{Weighting Quantities of the Reference Model.}
\begin{itemize}
\item $p_j(t)$: distribution of term $t$ across the $m$ concepts, normalized by its total frequency in the $SRD$; underlies discriminability (Equation~\ref{eq:term-concept-distribution}).
\item $\delta(t)\in[0,1]$: \emph{discriminability} of term $t$; ability of $t$ to discriminate among the concepts in the $SRD$, formulated from the normalized discrete entropy over the $m$ concepts (Equation~\ref{eq:discriminability}).
\item $n_j=|\mathcal{V}^{\mathrm{IND}}(R_j)|$: number of distinct independent terms in descriptor $R_j$ (Equation~\ref{eq:specificity-potentials}).
\item $a_\kappa(R_j)$: pattern-space size of type $\kappa$ for concept $c_j$; combinatorial for $\mathrm{IND}$, $\mathrm{2G}$, and $\mathrm{3G}$, and observed for $\mathrm{COC}$ (Equation~\ref{eq:specificity-potentials}).
\item \(\sigma(t\mid RTS_j)\in[0,1]\): \emph{specificity} of term $t$ with respect to concept $c_j$; quantifies the semantic information contributed by the term according to the restrictiveness of the linguistic pattern it represents and depends on its type $\kappa(t)$ (Equation~\ref{eq:specificity}).
\item $\sigma_{\mathrm{IND}}(R_j),\sigma_{\mathrm{2G}}(R_j),\sigma_{\mathrm{3G}}(R_j),\sigma_{\mathrm{COC}}(R_j)$: specificity values by term type for concept $c_j$ (Equation~\ref{eq:specificity}).
\item $\eta(t\mid RTS_j)\in[0,1]$: \emph{relevance} of term $t$ with respect to concept $c_j$; acts as a membership filter for the terminological profile, modulated by specificity (Equation~\ref{eq:relevance}).
\item $\varphi(t\mid RTS_j)\in[0,1]$: \emph{importance} of term $t$ for concept $c_j$, defined as the product of its intra-concept relevance and inter-concept discriminability (Equation~\ref{eq:importance}).
\end{itemize}

\paragraph{Reference Model.}
\begin{itemize}
\item $\mathbf{F}\in\mathbb{N}^{|\mathcal{V}_{SRD}|\times m}$: concept matrix, whose entry $F_{tj}=f(t\mid RTS_j)$ is the frequency of term $t$ in concept $c_j$ (Equation~\ref{eq:concept-matrix}).
\item $M$: reference model induced from the $SRD$; associates each term with its type and its importance per concept, together with the concept-specific reference affinities (Section~\ref{subsec:phase1}).
\end{itemize}

\paragraph{Affinity and Model Outputs.}
\begin{itemize}
\item $\ell(t;S_i,RTS_j)$: elementary affinity, or contribution, of term $t$ in unit $S_i$ to concept $c_j$ (Equation~\ref{eq:term-level-affinity}).
\item $\Lambda(D,SRD)\in\mathbb{R}^{n\times m}$: textual-unit--concept affinity matrix between the target document $D$ and the $SRD$ (Equation~\ref{eq:affinity-matrix}).
\item $\Lambda_{ij}(D,SRD)\in[0,1]$: component of $\Lambda(D,SRD)$ that quantifies the affinity of textual unit $S_i$ with respect to concept $c_j$ (Equation~\ref{eq:unit-level-affinity}).
\item $\mathcal{A}(\cdot)$: aggregation operator over the textual-unit axis (e.g., sum, mean, maximum, or percentiles), selected according to the analytical objective (Equation~\ref{eq:doc-level-affinity}).
\item $\lambda_j(D,SRD)$: aggregated absolute document-level affinity of document $D$ with respect to concept $c_j$ (Equation~\ref{eq:doc-level-affinity}).
\item $\lambda(D,SRD)\in\mathbb{R}^{m}$: document-level absolute affinity vector (Equation~\ref{eq:doc-affinity-vector}).
\item $\lambda^{\mathrm{ref}}_j(SRD)$: reference affinity of concept $c_j$; affinity of its own descriptor $R_j$ with respect to the profile $RTS_j$, which serves as an interpretive anchor (Equations~\ref{eq:ref-affinity}--\ref{eq:ref-affinity-expanded}).
\item $\lambda^{\mathrm{rel}}_j(D,SRD)$: relative affinity of document $D$ with respect to concept $c_j$, obtained by normalizing the absolute affinity by the reference affinity (Equation~\ref{eq:relative-affinity}).
\item $\lambda^{\mathrm{rel}}(D,SRD)\in\mathbb{R}^{m}$: document-level relative affinity vector (Equation~\ref{eq:relative-affinity-vector}).
\end{itemize}

\paragraph{Acronyms.}
\begin{itemize}
\item \textbf{ITL}: \emph{Intelligent Target Locator}; methodology proposed in this work.
\item \textbf{SRD}: \emph{Structured Reference Document}.
\item \textbf{RTS}: \emph{Reference Term Set} associated with a concept.
\item \textbf{TP}: \emph{Terminological-Profile Collection}.
\item \textbf{SDG}: \emph{Sustainable Development Goal}.
\item \textbf{NLP}: \emph{Natural Language Processing}.
\item \textbf{IND}, \textbf{2G}, \textbf{3G}, \textbf{COC}: term types considered (independent term, bigram, trigram, and co-occurrence, respectively).
\end{itemize}

\end{document}